\documentclass[3p,times]{elsarticle}
\usepackage[utf8]{inputenc} 
\usepackage[T1]{fontenc}    

\usepackage{graphicx}
\usepackage{subcaption}

\usepackage{amssymb} 

\usepackage{amsmath} 

\usepackage{lineno} 
\usepackage{hyperref} 

\usepackage{multirow} 

\modulolinenumbers[5] 

\begin{document}

\begin{frontmatter}

\title{SPI-BoTER: Error Compensation for Industrial Robots via Sparse Attention Masking and Hybrid Loss with Spatial-Physical Information }

\author[1]{Xuao Hou\fnref{fn1}} 
\author[2]{Yongquan Jia\fnref{fn1}}
\author[1]{Xiulei Wang}
\author[1,3]{Shijin Zhang\corref{cor1}}
\author[1,3]{Yuqiang Wu\corref{cor2}}

\affiliation[1]{organization={School of Software},
                addressline={Northwestern Polytechnical University}, 
                city={Xi'an},
                postcode={710129},
                state={Shaanxi},
                country={China}}
\affiliation[2]{organization={School of Computer Science and Technology},
                addressline={University of Science and Technology of China}, 
                city={Hefei},
                postcode={230026},
                state={Anhui},
                country={China}}                
\affiliation[3]{organization={Suzhou Key Lab of Aerospace Industry Software and Data Science},
                addressline={Yangtze River Delta Research Institute of NPU}, 
                city={Taicang},
                postcode={215400},
                state={Jiangsu},
                country={China}}

\fntext[fn1]{These authors contributed equally to this work.}          
\cortext[cor1]{Corresponding author. Email: zhangshijin@nwpu.edu.cn}
\cortext[cor2]{Corresponding author. Email: moran@nwpu.edu.cn}

\begin{abstract}
The widespread application of industrial robots in fields such as cutting and welding has imposed increasingly stringent requirements on the trajectory accuracy of end-effectors. However, current error compensation methods face several critical challenges, including overly simplified mechanism modeling, a lack of physical consistency in data-driven approaches, and substantial data requirements. These issues make it difficult to achieve both high accuracy and strong generalization simultaneously. To address these challenges, this paper proposes a Spatial-Physical Informed Attention Residual Network (SPI-BoTER). This method integrates the kinematic equations of the robotic manipulator with a Transformer architecture enhanced by sparse self-attention masks. A parameter-adaptive hybrid loss function incorporating spatial and physical information is employed to iteratively optimize the network during training, enabling high-precision error compensation under small-sample conditions. Additionally, inverse joint angle compensation is performed using a gradient descent-based optimization method. Experimental results on a small-sample dataset from a UR5 robotic arm (724 samples, with a train:test:validation split of 8:1:1) demonstrate the superior performance of the proposed method. It achieves a 3D absolute positioning error of 0.2515 mm with a standard deviation of 0.15 mm, representing a 35.16\% reduction in error compared to conventional deep neural network (DNN) methods. Furthermore, the inverse angle compensation algorithm converges to an accuracy of 0.01 mm within an average of 147 iterations. This study presents a solution that combines physical interpretability with data adaptability for high-precision control of industrial robots, offering promising potential for the reliable execution of precision tasks in intelligent manufacturing.

\end{abstract}






\end{frontmatter}



\section{Introduction}
\label{sec:1}
Driven by the wave of intelligent transformation under Industry 4.0, industrial robots, as a core enabling technology in advanced manufacturing systems, have continuously expanded their application scope from traditional discrete manufacturing to the domain of precision machining. In modern intelligent manufacturing systems, industrial robots are not only responsible for critical processes such as precision assembly \cite{1}, welding operations \cite{2}, and high-accuracy CNC machining\cite{3}, but also achieve groundbreaking applications in high-end manufacturing sectors like aerospace and automotive industries, where extremely stringent dimensional accuracy is required \cite{4,5}.However, during actual operations, the trajectory of the end effector often deviates from the theoretically planned path. Such deviations are mainly caused by factors including joint clearances, link deformations, and thermal drifts \cite{6,7}. The presence of these issues severely limits the performance of industrial robots in precision machining scenarios that demand high positional accuracy. Under current technological conditions, repeatability is generally superior to absolute positioning accuracy \cite{8,9}, and this performance discrepancy significantly hinders the deep integration and effective application of offline programming technologies in high-end manufacturing systems\cite{10}. Therefore, improving the absolute positioning accuracy of robots has long been a critical research topic in the field of intelligent manufacturing of high-end equipment. It is of profound and essential importance to advance the overall capabilities of the high-end manufacturing industry.
\begin{figure}
    \centering
    \includegraphics[width=0.5\linewidth]{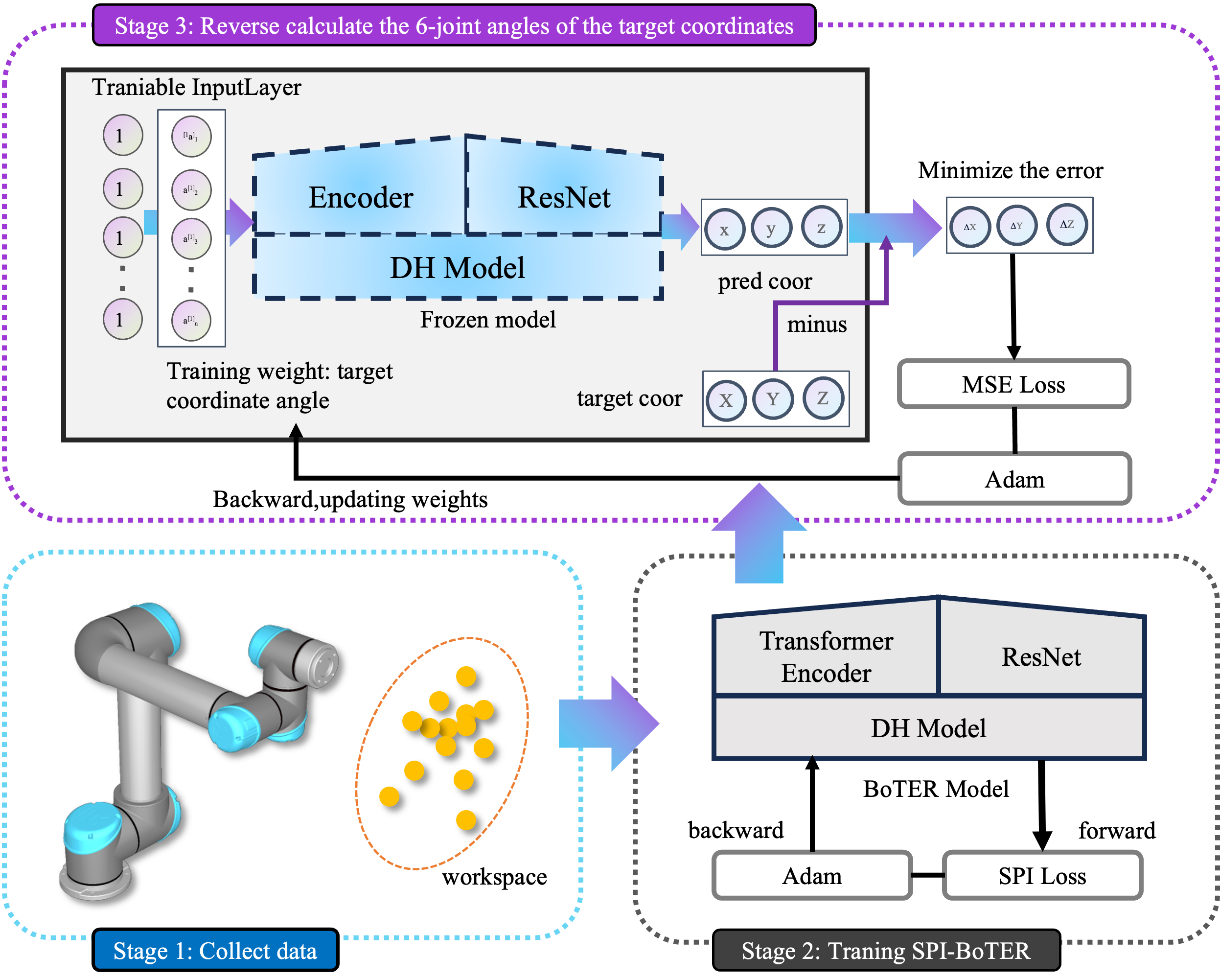}
    \caption{Schematic diagram of the complete error compensation process for industrial robots in this study}
    \label{fig:1.1}
\end{figure}

To ensure process quality, the end effector of a robotic manipulator must follow a high-precision trajectory. Conventionally, this is achieved by off-line programming for path planning in task space, followed by inverse kinematics to convert task-space instructions into joint-space motion parameters, and finally, forward kinematics is used to verify the accuracy of the planned path \cite{11}. The forward kinematics of a robot is typically modeled using Denavit–Hartenberg (D-H) parameters \cite{12}, which allow the theoretical position to be predicted through geometric modeling. This method describes the transformations between link coordinate systems using geometric parameters, allowing theoretical end-effector positions to be derived from joint angles. However, the D-H parameter method only considers link lengths, twist angles, offsets, and joint angles, and thus fails to capture the non-linear errors introduced during actual assembly. These unmodeled errors can accumulate, leading to absolute positioning deviations on the millimeter scale \cite{6}, severely hindering the execution of high-precision tasks.

Building upon the conventional Denavit–Hartenberg (D-H) method, improving D-H parameters has become a prominent direction in recent research, with many scholars proposing enhancements aimed at mitigating robotic positioning errors. Ibaraki Soichi et al. \cite{13} pointed out that thermal expansion of robotic joints significantly increases 2D positioning errors in planar manipulators. They demonstrated that accurately identifying the D-H parameters, specifically link lengths and rotational axis offsets, can effectively compensate for thermal effects, thereby reducing positioning errors. Yang Ping’s team \cite{14} developed a spatial error measurement device for a serial robot of six axes with a rotational dual-sphere structure. Using a modified D-H model and a virtual joint method to establish an error propagation model, combined with the Newton-Raphson compensation algorithm and ridge regression correction, they improved positioning accuracy from 0.0163 to 0.0236 mm in a load range of 1.0 to 2.0 kg. Feng Xinyu et al. \cite{15} made a breakthrough by integrating errors of the DH parameter with axis inclination errors, constructing a multi-source error coupling model for collaborative robots. They established, for the first time, a mapping relationship between joint axis inclination in the calibration coordinate system and end-effector errors, partially overcoming the modeling limitations of traditional nongeometric error compensation methods and enhancing positioning reliability in collaborative operation scenarios. Cui Tianhao’s team\cite{16} proposed a posture-dependent DH error model that introduced dynamic correction coefficients based on joint angles, allowing adaptive compensation of positioning errors. Furthermore, Wang Zhirong et al. (2019) \cite{17} developed a two-stage parameter identification approach that jointly identifies dynamic and geometric errors. This method effectively addresses joint deformation and load-dependent errors under heavy-load conditions, with its positioning accuracy improvements validated through experimental tests.

Although the aforementioned methods have made significant progress in improving robot positioning accuracy, they share a common strength in their ability to achieve coordinated optimization of geometric and dynamic errors through multi-source error coupling modeling and compensation strategies. However, current technologies still face key challenges, including high model complexity, substantial computational demands, and a strong reliance on extensive training data. In particular, their real-time responsiveness to nonlinear errors under dynamic operating conditions remains insufficient and requires further enhancement.

With the continuous advancement of artificial intelligence, data-driven approaches such as deep neural networks have introduced new perspectives for error compensation and have demonstrated unique advantages in the field of kinematic modeling. These methods can effectively capture nonlinear error characteristics that are difficult to describe using traditional models \cite{18}. Zhu Xin et al. \cite{19} proposed an interpretable prediction and compensation method based on deep learning, which significantly improved the pose accuracy of parallel robots. In particular, they were the first to apply the SHapley Additive Explanations (SHAP) method to interpret the factors that contribute to the predicted pose deviations, thereby enhancing the interpretability of deep learning models in this domain. As a result, the mean position and orientation deviations were reduced by 92.21\% and 89.45\%, respectively. Jin Zujin et al. \cite{20} developed an end effector trajectory error prediction model based on a Bayesian long-short-term memory network (BO-LSTM), which substantially decreased prediction errors for large optical mirror processing robots (LOMPR), thus improving both predictive accuracy and efficiency, and laying a foundation for enhancing surface precision in optical mirror machining.

Bucinskas Vytautas et al. \cite{21} applied an online deep Q-learning approach to analyze experimentally predefined robotic characteristics and their impact on overall precision. This method significantly improved the location accuracy at the key points of the trajectory, resulting in a reduction of more than 30\% in the position error of the key points, greatly enhancing the operational precision of the robots. Kato Daiki et al. \cite{22} constructed an offline teaching system based on 3D computer-aided design data and used convolutional neural networks (CNN) to predict robot positioning errors. Their model successfully identified critical features such as reverse teaching of joint angles at motion initiation, current overshoot in rotational joints, and current variations in swinging joints, which substantially improved both the motion accuracy of industrial robots and the effectiveness of compensation strategies.

In addition, breakthrough progress has been achieved through techniques such as the ADANN-2R transfer learning method \cite{23}, deep belief network (DBN) based optimization models [24], and hybrid kinematic-neural network fusion frameworks \cite{25}. However, it is worth noting that purely data-driven models exhibit a "black-box" nature. Their predictions may violate the underlying physical laws, which limits their applicability in scenarios requiring high reliability. Moreover, these models typically demand thousands of training samples to ensure stability and generalization \cite{25}.

In recent years, physics-informed data-driven approaches have opened new avenues for research in error compensation. By embedding physical mechanism models into the architecture or loss function of neural networks, these methods preserve the powerful nonlinear learning capabilities of data-driven models while ensuring that the predictions adhere to physical constraints \cite{26}. Physics-Informed Neural Networks (PINNs) incorporate kinematic equation constraints into the loss function, enabling the model to fit experimental data while satisfying physical laws. This not only reduces overfitting, but also allows high performance with relatively few samples \cite{27}.Li Wenjing et al. \cite{28} developed an Embedded Approximate Function Boundary Condition PINN (BCAF-PINN), which innovatively encodes initial and boundary conditions directly into the approximate function. This approach successfully avoids the erroneous convergence caused by gradient imbalance in traditional PINNs when solving forward and inverse problems in flexible electromechanical systems and soft robots. It significantly reduced parameter identification errors and revealed physical characteristics during joint flexion and extension.Hongbo Hu’s team \cite{29} proposed a friction-inclusive dynamic PINN modeling method tailored for industrial robots. By integrating the Lagrangian dynamics framework with the Stribeck friction model, they achieved an average reduction of 39.69\% in joint torque errors, demonstrating the strong generalizability of the method in complex dynamic environments. Xingyu Yang et al. \cite{30} introduced a hybrid modeling framework based on a Physics-Informed Neural Network (H-PINN). By embedding the physical laws of joint dynamics into a recurrent neural network (RNN) architecture, they achieved highly accurate predictions and parameter identification of collaborative robot joint dynamics. In predicting the dynamic response of the collaborative robot joints UR3e, this method reduced the standard deviation of positioning error by 39.69\% compared to baseline methods, validating its superior performance in complex non-linear systems. These approaches effectively balance physical interpretability and data-driven flexibility, achieving a synergistic optimization of modeling accuracy and computational efficiency in robotic error compensation tasks.

In terms of innovation in model architecture, the introduction of Transformer and attention mechanisms has provided breakthrough tools to extract complex error features \cite{31}. The self-aware spatiotemporal feature fusion method allows dynamic identification of the time-varying influence weights of different error sources, such as joint backlash and thermal drift, on the end effector pose. Chen Jinlong et al. \cite{32} implemented a Transformer-based model capable of directly regressing translational and rotational errors between the end effector of a robot and the target object, thus significantly improving the efficiency of robotic pollination tasks. Li Feng et al. \cite{33} proposed the DDETR-SLAM algorithm, which introduces a deformable detection transformer (DETR) network and incorporates semantic information. This effectively reduces localization errors in dynamic environments and significantly enhances the accuracy and robustness of SLAM. In highly dynamic scenarios, the algorithm achieved reductions in absolute trajectory error, translational error, and rotational error by 98. 45\%, 95. 34\%, and 92. 67\%, respectively, compared to ORB-SLAM2. Compared with traditional CNN-LSTM \cite{34} architectures, Transformer demonstrates a 36\% reduction in memory decay when modeling long-sequence error propagation \cite{35}, which makes it particularly well suited for dynamic compensation of cumulative errors under complex operating conditions.

Although Transformer models have demonstrated outstanding performance in sequence modeling due to their self-attention mechanism, the conventional Transformer architecture does not consider the kinematic characteristics of six-axis serial robotic arms. As a result, attention weights may be distributed across irrelevant joints. The Body Transformer addresses this issue by incorporating structural information from the robot body to optimize policy learning \cite{36}. Its sparse attention mechanism based on physical connectivity offers key insights for this study.

Inspired by the hierarchical attention design of the iTransformer \cite{37} and the sparse self-attention masking strategy of the Body Transformer \cite{36}, this study proposes the Spatial Physical Information Attention Network (SPI-BoTER), which achieves high-precision error compensation through three innovative architectural components:
\begin{enumerate}
    \item Dual-branch mechanism-data collaborative network: This design decouples theoretical coordinate prediction based on the DH model from Transformer-driven error compensation, thereby avoiding the limitations of a single modeling paradigm.
    \item Sparse self-attention masking tailored to six-axis serial manipulator kinematics: By constraining inter-joint interaction relationships, this mechanism improves feature extraction efficiency. Additionally, residual networks are integrated to enhance the stability of the compensation value prediction.
	\item Hybrid spatial physical information loss function: This function aligns the Euclidean distance matrices between theoretical models and prediction outputs, enforcing spatial topological consistency of robotic motion. A dynamic weighting mechanism is further introduced to balance multi-objective optimization processes.
\end{enumerate}
By integrating kinematic constraints with the flexibility of data-driven modeling, the proposed framework achieves high-precision error compensation when trained on the UR5 robot dataset. This provides a solution for six-axis serial manipulator error compensation that balances physical interpretability with modeling flexibility.

Against the backdrop of industrial robots increasingly penetrating high-precision manufacturing scenarios, error compensation technology is undergoing a transformative shift, from traditional mechanism-based modeling to integrated intelligent learning paradigms. However, key challenges remain, including insufficient modeling of non-linear errors, inadequate guarantees of physical consistency, and limited adaptability to complex working conditions. The Spatial Physical Information Attention Network (SPI-BoTER) proposed in this study is an innovative attempt driven by these demands. It aims to develop an error compensation model that combines high-precision prediction with strong physical constraints, achieved through structurally aware sparse attention mechanisms and a physics-guided hybrid loss function.

The main contributions of this paper include:
\begin{itemize}
	\item[(1)] \textbf{A dual-branch Transformer architecture tailored for six-axis industrial robots}: This is the first approach to integrate sparse self-attention mechanisms with the motion chain structure of robotic arms, enabling efficient feature extraction.
	\item[(2)] \textbf{A parameter-adaptive hybrid loss function incorporating spatial physical information}: This design overcomes the limitations of traditional Physics-Informed Neural Networks (PINNs) that rely on partial differential equations, enhancing physical consistency through Euclidean distance matrix constraints.
	\item[(3)] \textbf{A gradient descent-based inverse joint angle compensation algorithm}: This algorithm provides high-precision correction inputs for closed-loop control systems.
	\item[(4)] \textbf{Comprehensive validation on an industrial-grade robotic platform}: The proposed method demonstrates superior performance in terms of error compensation accuracy, training efficiency, and generalizability.
\end{itemize}
These research results offer both theoretical support and a technical framework for achieving high-precision robotic control in intelligent manufacturing environments.

\section{Methodology}
\label{sec:2}
To address the tasks of high precision point prediction for robotic arms and inverse joint angle error compensation, this study proposes the use of the \textbf{Body-Transformer Enhanced Residual Network (BoTER) }optimized with a \textbf{Spatial Physical Information Hybrid Loss} for accurate point prediction. Meanwhile, the inverse solution of joint angles for industrial robots is formulated as a multi-objective optimization problem. Based on the best-performing trained model, a gradient descent backpropagation algorithm using the Adam optimizer [38] is used to optimize the input joint angles and obtain the corresponding angle compensation values. The details are elaborated as follows:
\subsection{BoTER Architecture}
\label{sec:2.1} 
BoTER is a dual stream architecture that fuses physical principles and data-driven learning, designed for the prediction of end effector positions in six-axis industrial robots. By integrating the prior kinematic knowledge of robotic arms with the nonlinear representation capabilities of deep learning, it achieves high-precision prediction of end-effector positions.
\begin{figure}[ht]
    \centering
    \includegraphics[width=0.75\linewidth]{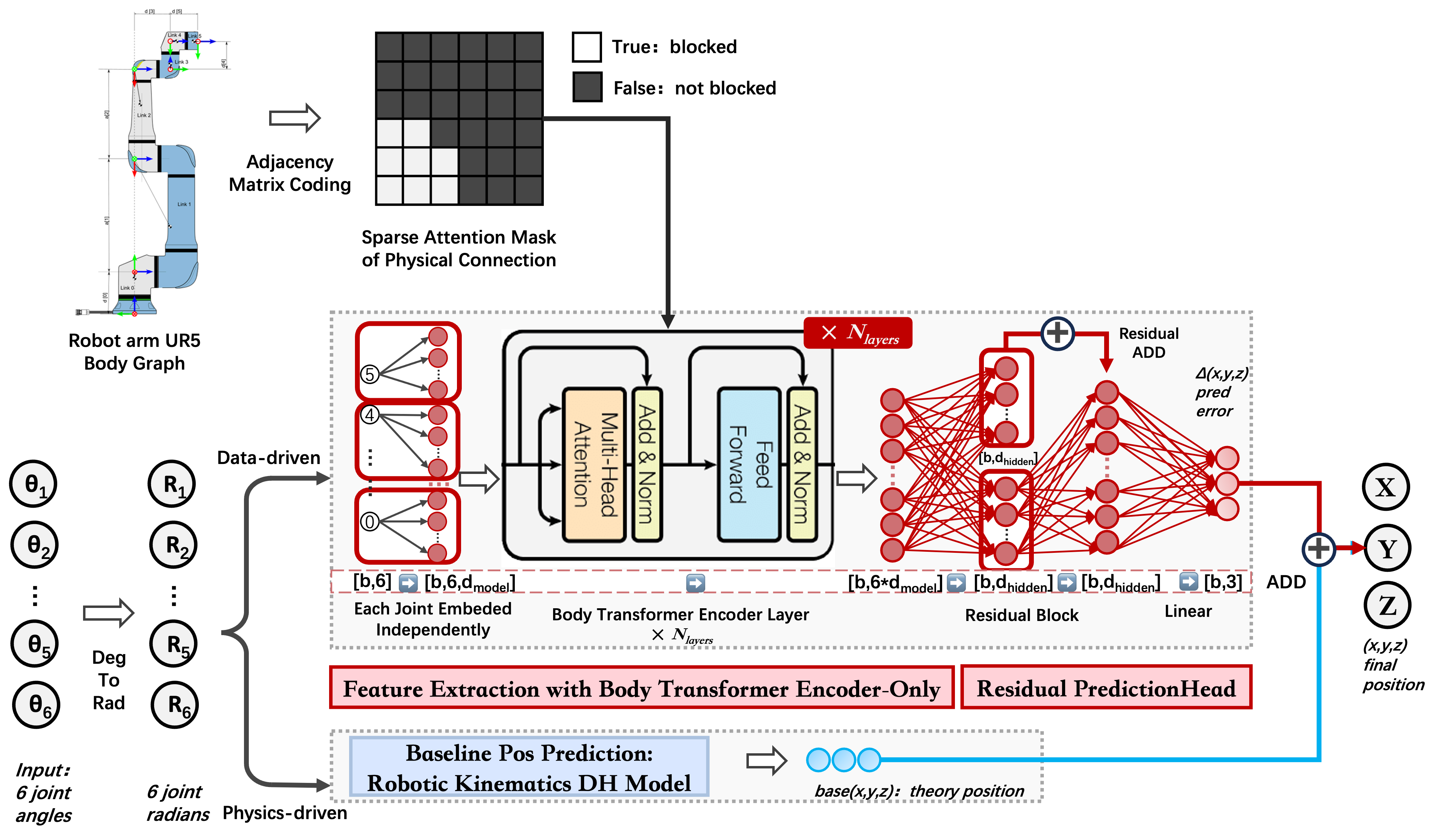}
    \caption{ Schematic Diagram of the BoTER Model Architecture}
    \label{fig2.1}
\end{figure}

As shown in \ref{fig2.1}, the internal structure of BoTER consists of two coordinated branches to process joint angles: a physics-driven branch and a data-driven branch. Let $\theta = [\theta_1, …, \theta_6]$ represent the six joint angles of the robotic arm. These angles are first converted into radians, denoted as $R = [R_1, …, R_6]$, and then inputted into the BoTER network. The output of the physics-driven branch, $base[x, y, z]$, corresponds to the prediction of the robot's forward kinematics. The data-driven branch predicts the coordinate compensation values $\Delta[x, y, z]$. The final predicted position $[X, Y, Z]$ is obtained by combining both outputs.

\subsubsection{Forward Branch 1: Physik-Driven Robotic Arm Kinematic Model}
\label{sec:2.1.1}

This branch constructs a kinematic physics sub-network based on the Denavit-Hartenberg (DH) parameterization method. It takes a 6-dimensional joint angle vector $\boldsymbol{\theta} \in \mathbb{R}^6$ as input and outputs the theoretical position of the end effector $\mathbf{Y}_{\text{theory}} \in \mathbb{R}^3$. The detailed computation steps are as follows:

\begin{itemize}
\item[(1)] \textbf{Load DH Parameters}

At initialization, a differentiable tensor is generated from the preset DH parameter table $[d, \theta_{\text{offset}}, a, \alpha]$, which includes link offset $d$, joint angle offset $\theta_{\text{offset}}$, link length $a$, and twist angle $\alpha$. This tensor is configured to be compatible with GPU/CPU computation.

\item[(2)] \textbf{Coordinate Transformation for Each Joint}

For each joint angle $\theta_i$, the corresponding homogeneous transformation matrix $T_i \in \mathbb{R}^{4 \times 4}$ is computed using the following equation\ref{equ:1}:
\begin{equation}
T_i =
\begin{bmatrix}
\cos\theta_i & -\sin\theta_i \cos\alpha_i & \sin\theta_i \sin\alpha_i & a_i \cos\theta_i \\
\sin\theta_i & \cos\theta_i \cos\alpha_i & -\cos\theta_i \sin\alpha_i & a_i \sin\theta_i \\
0 & \sin\alpha_i & \cos\alpha_i & d_i \\
0 & 0 & 0 & 1
\end{bmatrix}
\tag{1}
\label{equ:1}
\end{equation}

\item[(3)] \textbf{Cumulative Transformation Matrix}

By sequentially multiplying the individual transformation matrices in the order of the robotic arm’s links, the final homogeneous transformation matrix $T_{\text{end}}$ is obtained, representing the transformation from the base frame to the end-effector frame.

\item[(4)] \textbf{Tool Frame Compensation}

The transformation matrix is further multiplied by the external tool offset matrix. The translational component of the resulting matrix is extracted as the theoretical end-effector position:$\mathbf{Y}_{\text{theory}} = T_{\text{end}}[:3, 3]$.
This value serves as the output of the physics-driven kinematic model.

\end{itemize}

\begin{table}
    \centering
    \begin{tabular}{ccccc}
      \hline
      Joint & $d$($mm$) & $a$($mm$) & $\alpha$($rad$) & $\theta$($rad$)\\
      \hline
      1 & 89.159 & 0       & $\pi/2$  & 0 \\
      2 & 0      & -425    & 0               & 0\\
      3 & 0      & -392.25 & 0               & 0\\
      4 & 109.15 & 0       & $\pi/2$  & 0\\
      5 & 94.65  & 0       & $-\pi/2$ & 0\\
      6 & 82.3   & 0       & 0               & 0\\
      \hline
    \end{tabular}
    \caption{DH Parameters of the UR5 Robotic Arm}
    \label{tab:my_label}
\end{table}
\subsubsection{Feedforward Branch 2: BoTER Data Driven Module}
\label{sec:2.2.1}
Building on practical observations and considering the nonlinear characteristics of multi-joint coupling errors in six-axis industrial robots, we modify the Body Transformer Encoder architecture proposed by \cite{36} to enhance feature extraction. In addition, a residual network is introduced at the model head to perform the final prediction.

Since its proposal by Vaswani et al. \cite{31}, the Transformer architecture has achieved breakthroughs in fields such as machine translation, time series prediction, and computer vision. Its ability to model global dependencies makes it particularly well suited for error compensation tasks in high-dimensional nonlinear systems like industrial robots.

In this study, we apply a modified Transformer architecture to joint error compensation in six-axis robots for the first time. By visualizing the attention weights, we can interpret the error propagation paths, offering both high prediction accuracy and model interpretability.

The key aspects are as follows.
\begin{itemize}
\item[(1)] \textbf{Input Token Processing}

The joint angles $\theta = [\theta_1, \ldots, \theta_6]$ are first converted to radians $R = [R_1, \ldots, R_6]$. Considering that the joint angle errors of a six-axis manipulator share the same unit and measurement range (that is,$\theta \in [-\pi, \pi]$), we omit normalization for the dataset.

Traditional Transformer models embed multiple variables at each time step in a single token, which can result in meaningless attention maps and severely weaken the model's ability to learn from time sequences \cite{37}. To align with the input dimension of the Transformer Encoder $d_{\text{model}}$, and following the high-dimensional feature decoupling approach from \cite{36, 37}, we project each joint angle value $R_i$ into a high-dimensional feature space via an independent linear layer (from 6 to $d_\text{model}$).

The embedded feature representation $F_{\text{embed}} \in \mathbb{R}^{\text{batch size} \times 6 \times d_{\text{model}}}$ is computed as shown in Equation 
\ref{equ:2},\ref{equ:3}:
\begin{equation}
F_{\text{embed}}^i = \text{ReLU}(\mathbf{W}_{\text{embed}} \mathbf{R}_i + \mathbf{b}_{\text{embed}}), \quad \mathbf{W}_{\text{embed}}, \mathbf{b}_{\text{embed}} \in \mathbb{R}^{d_{\text{model}} \times 1} 
\tag{2}
\label{equ:2}
\end{equation}

\[
F_{\text{embed}} = [F_{\text{embed}}^0, \ldots, F_{\text{embed}}^i, \ldots, F_{\text{embed}}^5], \quad i = 0, \ldots, 5 \tag{3} \label{equ:3}
\]

In addition, since the end-effector position data of the UR5 industrial robot used in our experiments are static and randomly sampled at the same time step, the positional encoding layer is removed as in \cite{36}.

\item[(2)] \textbf{Design of Sparse Self-Attention Mask Adapted to the Six-Axis Serial Industrial Robot Structure} 

Consistent with the structural characteristics of the Six-Axis serial industrial robots, we designed a Sparse Self-Attention Mask Adapted to this configuration following the physical linkage rules and kinematic properties of the manipulator \cite{38}.

Taking the serial six-axis industrial robot UR (as shown in Figure \ref{fig:2.2}) as an example, its kinematic structure features a hierarchical division of functions: the first three joints primarily handle coarse position adjustment, while the latter three joints are responsible for fine orientation refinement.

\begin{figure}[ht]
    \centering
    \includegraphics[width=0.25\linewidth]{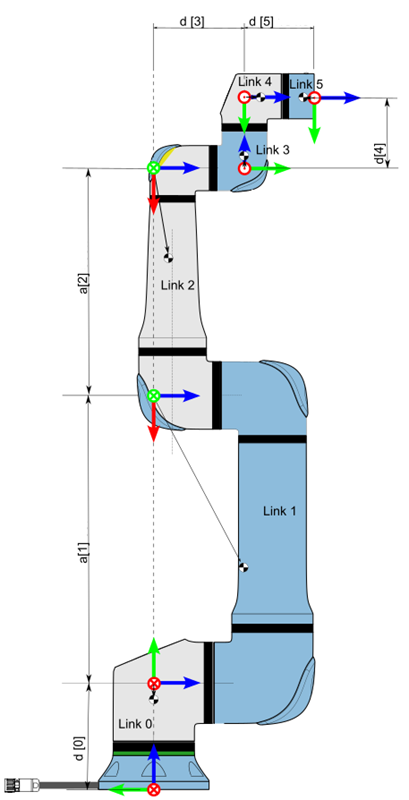}
    \caption{Kinematics properties for UR robot}
    \label{fig:2.2}
\end{figure}

We allow each joint to attend to its directly connected parent/child joints—i.e., neighboring joints are mutually visible. However, in contrast to the masking rules in Body Transformer [36], we design a hierarchical attention structure based on the functional characteristics of serial six-axis industrial robots. Specifically, the first three joints are grouped for coarse position control, while the latter three joints are grouped for fine orientation control.

Within each group, the joints are fully visible to one another. To preserve the directionality of the kinematic chain, we introduce unidirectional cross-group visibility: the first three joints can attend to the latter three, but not vice versa.

The final attention mask M is defined in Equation \ref{equ:4}:

\[
M = \begin{bmatrix}
0 & 0 & 0 & 0 & 0 & 0 \\
0 & 0 & 0 & 0 & 0 & 0 \\
0 & 0 & 0 & 0 & 0 & 0 \\
1 & 1 & 0 & 0 & 0 & 0 \\
1 & 1 & 1 & 0 & 0 & 0 \\
1 & 1 & 1 & 0 & 0 & 0 \\
\end{bmatrix}
\quad \text{where } 0 = \text{visible},\ 1 = \text{masked}
\tag{4}
\label{equ:4}
\]

Here, \textbf{0} indicates \emph{visible} (i.e., attention is allowed), while \textbf{1} denotes \emph{masked} (i.e., attention is blocked).

\item[(3)] \textbf{Global Feature Extraction via Transformer Encoder Layers}

The sparse attention mask derived in the previous step is passed as the padding mask into the Transformer encoder layer. We employ $N_{\text{layer}}$ standard Transformer encoder layers based on the PyTorch implementation, where the input feature dimension is $d_{\text{model}}$, and the feedforward layer has a hidden size of $4 \times d_{\text{model}}$. Each layer consists of $n_{\text{head}}$ self-attention heads and a feedforward subnetwork. The self-attention mechanism is used to compute the global dependency relationships among joint angles \cite{36}, as shown in Equation \ref{equ:5},\ref{equ:6}:

\[
\text{AttenScores} = \frac{QK^\top}{\sqrt{d_k}}, \quad \text{where } QK^\top \in \mathbb{R}^{n \times n}
\tag{5}
\label{equ:5}
\]

\[
\text{MaskedAttenScores} = \text{AttenScores} + M
\tag{6}\label{equ:6}
\]

where \( M \) is the attention mask.

\item[(3)] \textbf{Output Layer Design for Error Compensation Prediction}

The next step involves an output layer composed of a two-layer residual block network, where the ReLU activation function \cite{40} is used to enhance nonlinearity. Specifically, the feature vector passes through a residual block that projects from $d_{\text{model}} \times 6$ to a hidden dimension $d_{\text{hidden}}$, followed by a fully connected linear layer from $d_{\text{hidden}}$ to the output dimension of 3. The final compensation output is computed as shown in Equation \ref{equ:7}:

\[
\Delta p = \text{Linear}_3 \left( \text{ReLU} \left( \text{ResidualBlock}(x) \right) \right)
\tag{7}
\label{equ:7}
\]

The residual block structure used in the head of the BoTER model is detailed as shown in Equation \ref{equ:8}:

\[
\begin{aligned}
a &= \text{ReLU} \left( \text{Linear}_1(x) \right) \\
b &= \text{Linear}_2(a) \\
r &= \text{Linear}_s(x) \\
z &= \text{ReLU}(b + r)
\end{aligned}
\tag{8}
\label{equ:8}
\]

\end{itemize}

\subsubsection{Fusion Mechanism}
\label{sec:2.1.3}

After passing through the two branches, the input joint angles of the six-axis robotic arm are fused by combining the theoretical kinematic prediction and the learned compensation from the neural network. The final predicted end effector position is calculated as shown in \ref{equ:9}:

\[
p_{\text{pred}} = p_{\text{theory}} + \alpha \cdot \Delta p
\tag{9}
\label{equ:9}
\]

Here, $\alpha$ is initialized to $0.1$ and is defined as a trainable scaling parameter to ensure stable convergence and optimal fusion performance.

\subsection{SPI Loss Function Design}
\label{sec:2.2}
This study proposes the spatially-physically informed (SPI) hybrid loss function, which integrates kinematic priors with data-driven features to achieve high-precision compensation while preserving physical consistency. The design of the SPI loss function follows three core principles:
1.	Physical Interpretability: The differential constraints of classical kinematics remain.
2. Data Adaptability: It captures the statistical patterns of the unmodeled errors.
3. Dynamic balance: It autoadjusts the optimization weights between physical constraints and data fitting.

\subsubsection{Data Residual Term}
\label{sec:2.2.1}

This term serves as the fundamental optimization component of the loss function. It ensures alignment between the predicted and actual positions of the end-effector. It is defined as the geometric error between the prediction of the model and the truth of the ground, as shown in Equation \ref{equ:10}:

\[
L_{\text{data}} = \frac{1}{N} \sum_{i=1}^{N} \left\| p_{\text{pred}}^{(i)} - p_{\text{gt}}^{(i)} \right\|_2^2
\tag{10}
\label{equ:10}
\]

where $p \in \mathbb{R}^{3 \times 6}$ represents the position vector of the end effector.

\subsubsection{Spatial Physical Information Residual Term}
\label{sec:2.2.2}

\begin{figure}[ht]
    \centering
    \includegraphics[width=0.75\linewidth]{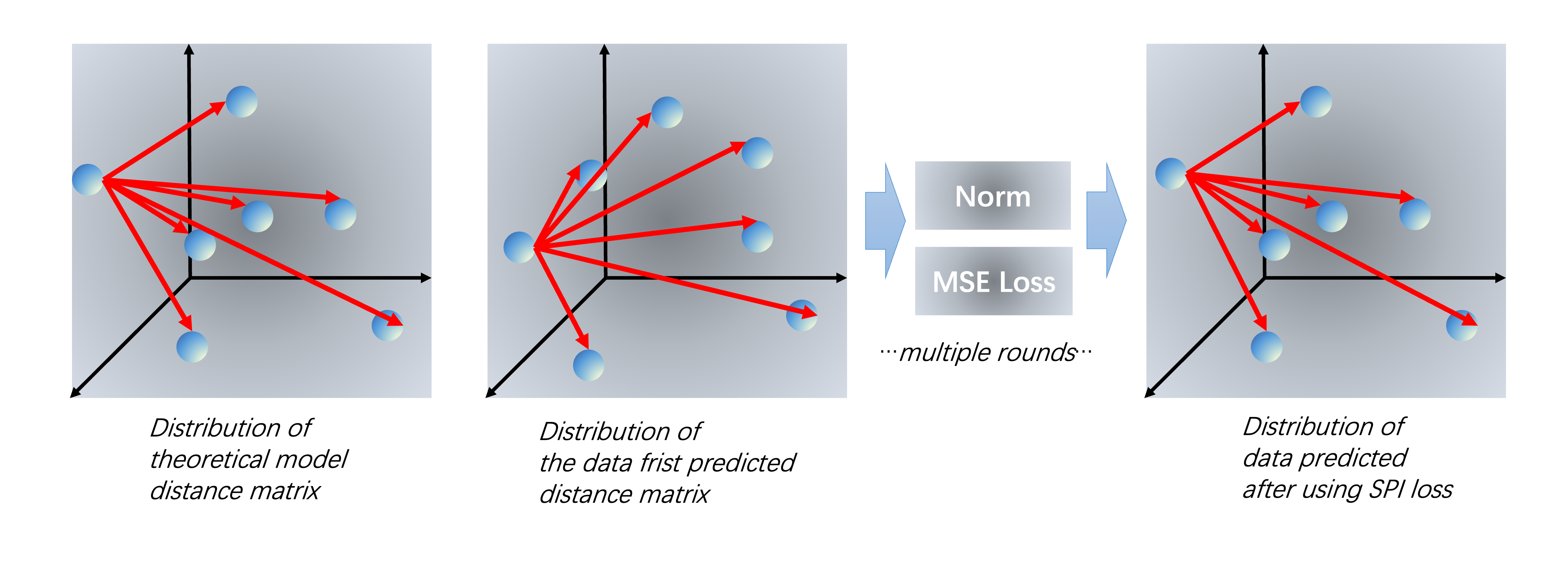}
    \caption{Simulation of one batch of training samples optimized using the SPI loss function.}
    \label{fig:2.3}
\end{figure}

We introduce a novel mechanism for the preservation of spatial structure leveraging the Euclidean distance matrix \cite{41} as a representation of the spatial topology. The spatial structure generated by the theoretical kinematic model is used to guide the distribution of the predicted outputs from the data-driven model (as illustrated in Fig \ref{fig:2.3}). This approach significantly improves the geometric consistency of the robotic arm point prediction. Enhance physical interpretability while mitigating the degradation in prediction performance that may occur when the discrepancy between the theoretical and measured values is large.
\begin{figure}[htbp]
    \centering
    \begin{subfigure}[b]{0.3\textwidth}
        \includegraphics[width=\textwidth]{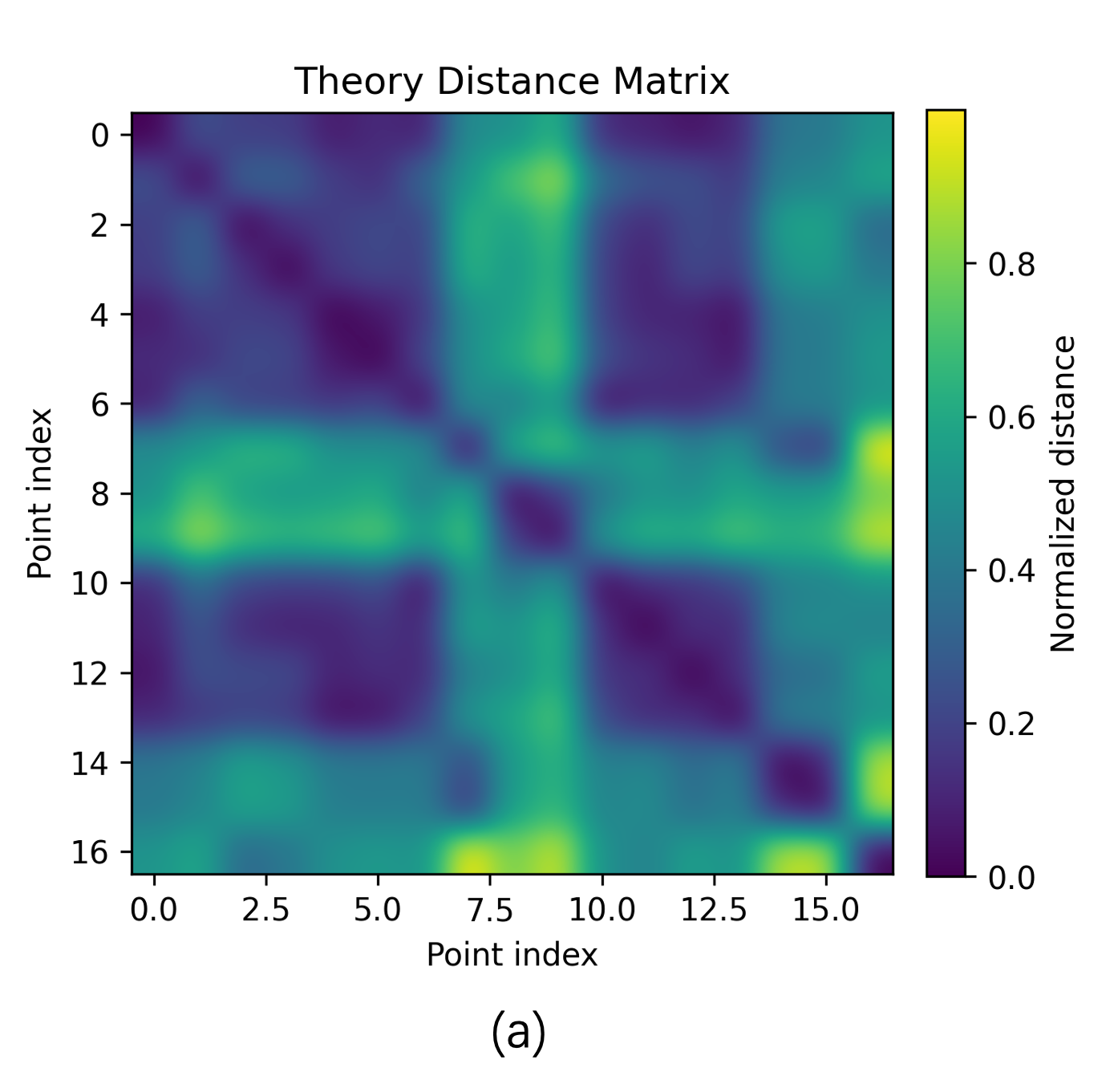}
        \caption{Theoretical distance matrix of a batch of test samples.}
        \label{fig:2_4a}
    \end{subfigure}
    \hfill
    \begin{subfigure}[b]{0.3\textwidth}
        \includegraphics[width=\textwidth]{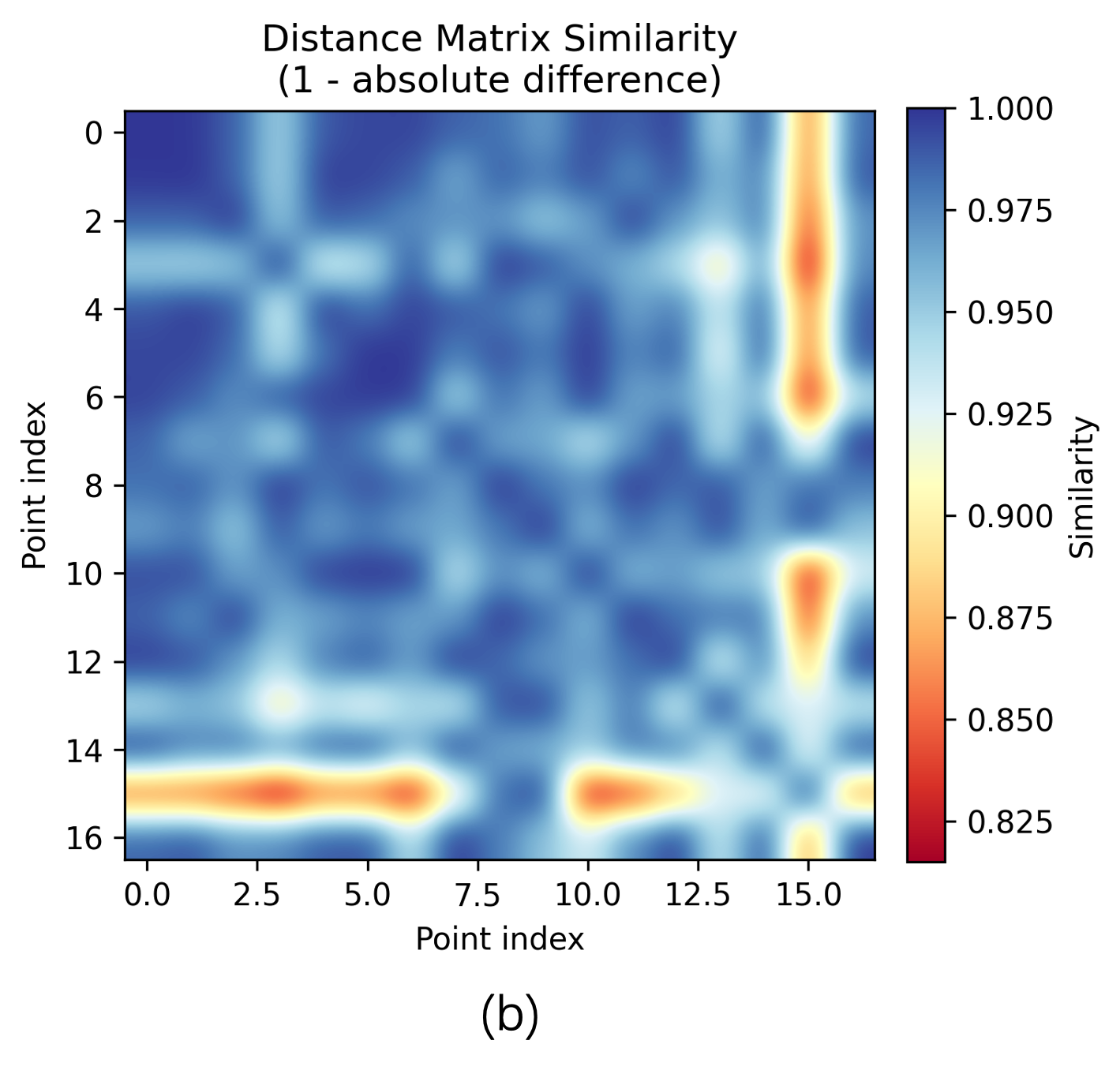}
        \caption{Predicted distance matrix of the same batch of test samples.}
        \label{fig:2_4b}
    \end{subfigure}
    \hfill
    \begin{subfigure}[b]{0.3\textwidth}
        \includegraphics[width=\textwidth]{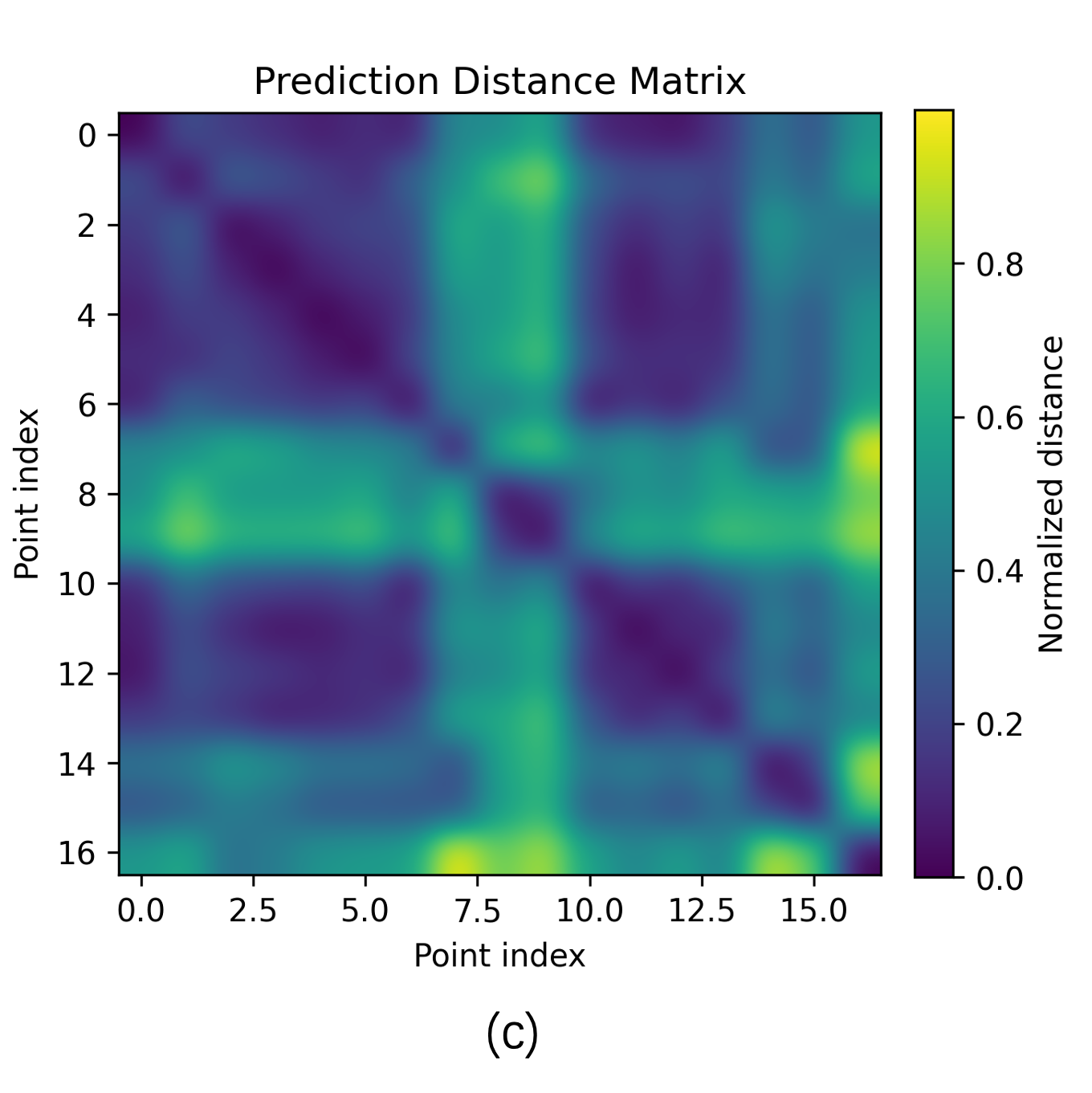}
        \caption{Overlay of the theoretical and predicted distance matrices.}
        \label{fig:2_4c}
    \end{subfigure}
    \caption{Comparison of distance matrices for a batch of test samples.}
    \label{fig:2_4}
\end{figure}

The theoretical distance matrix \( D_{\text{theory}} \in \mathbb{R}^{N \times N} \) is calculated based on the theoretical coordinates derived from the Denavit-Hartenberg (D-H) model. For each training batch containing \( N \) robot arm poses, the computation process is as follows:

\begin{enumerate}[(1)]
    \item The coordinates of the end effector are computed using the modified Denavit-Hartenberg model: \begin{equation}
        Y_{\text{theory}}^i = f_{\text{DH}}(\theta_i), \quad i = 1, \dots, N
        \label{eq:theory_position}\tag{11}
    \end{equation}
    where \( \theta_i \in \mathbb{R}^6 \) denotes the joint angle vector and \( f_{\text{DH}} \) represents the improved Denavit-Hartenberg kinematic model.

    \item The Euclidean distance matrix is then constructed: \begin{equation}
        D_{\text{theory}}[i, j] = \| Y_{\text{theory}}^i - Y_{\text{theory}}^j \|^2
        \label{eq:theory_distance}\tag{12}
    \end{equation}
    This matrix encodes the intrinsic geometric relationships among the points generated by the theoretical model within each training batch. As illustrated in the left subfigure of Fig.~\ref{fig:2_4a}, the theoretical distance matrix exhibits a distinct blockwise structure, reflecting spatial clustering patterns along the continuous motion trajectory of the robotic arm.

    \item The predicted distance matrix \( D_{\text{pred}} \in \mathbb{R}^{N \times N} \) is calculated based on the predicted positions from the BoTER model: \begin{equation}
        D_{\text{pred}}[i, j] = \| y_{\text{pred}}^i - y_{\text{pred}}^j \|^2
        \label{eq:pred_distance}\tag{13}
    \end{equation}

    \item To eliminate absolute scale variations and focus on preserving relative spatial relationships, both matrices are normalized: \begin{equation}
        D_{\text{theory-norm}} = \frac{D_{\text{theory}}}{\max(D_{\text{theory}})}
        \label{eq:norm_theory}\tag{14}
    \end{equation}
    \begin{equation}
        D_{\text{pred-norm}} = \frac{D_{\text{pred}}}{\max(D_{\text{pred}})}
        \label{eq:norm_pred}\tag{15}
    \end{equation}

    \item The spatial physics-informed loss term is defined as follows: \begin{equation}
        L_{\text{physics}} = \frac{1}{N^2} \sum_{i,j} \left\| D_{\text{pred-norm}}[i,j] - D_{\text{theory-norm}}[i,j] \right\|^2
        \label{eq:physics_loss}\tag{16}
    \end{equation}
\end{enumerate}

This loss function calculates the mean squared error (MSE) between the normalized predicted and theoretical distance matrices across each training batch, thereby enforcing the predicted keypoints to maintain a spatial topology consistent with the theoretical model.

Fig.\ref{fig:2_4} presents the structural comparison between the relative distance matrices generated by the theoretical and learned models on the test set using the final trained BoTER model. Aside from an outlier at sample 15, the heatmap displays a uniformly blue color, indicating a high degree of spatial similarity between the predicted and theoretical structures.

\subsubsection{Dynamic Weight Adjustment Mechanism}
\label{sec:2.2.3}

To balance the importance of each loss term and adaptively search for the optimal weight combination, the SPI loss introduces a set of trainable parameters $\lambda_{\text{data}}$ and $\lambda_{\text{physics}}$. In practice, the implementation includes the following strategies:

\begin{itemize}
\item To prevent the weights from taking negative values, logarithmic parameterization is adopted:
\begin{equation}
\lambda^* = \exp(\log \lambda^*)
\label{eq:lambda_exp}\tag{17}
\end{equation}

\item The total loss is formulated as the weighted sum of the individual loss terms:
\begin{equation}
L_{\text{total}} = \lambda_{\text{data}} L_{\text{data}} + \lambda_{\text{physics}} L_{\text{physics}}
\label{eq:total_loss}\tag{18}
\end{equation}
\end{itemize}

As training epochs progress, these weight parameters are automatically updated through backpropagation, enabling a dynamic adjustment of the relative importance between data fitting and physics-informed constraints.

\subsection{Gradient-Based Inverse Angle Compensation Algorithm Using the Trained BoTER Model}
\label{sec:2.3}

Another major objective of robotic arm error compensation is to infer joint angle corrections from a desired end-effector position, i.e., solving the inverse problem. Traditional numerical approaches have limited capability in handling the non-linear nature of such compensation tasks.

In this work, we formulate the problem as a multi-objective optimization task, where the goal is to minimize the residual errors along the x, y, and z dimensions, aiming toward a zero tensor as the optimization target. Gradient descent, as a classic algorithm for multi-objective optimization, offers advantages in both convergence speed and accuracy\cite{38}.

To this end, we freeze the parameters of the trained BoTER model and treat it as a forward-kinematics solver. A trainable input layer is introduced\cite{38}, and the optimization is carried out using the Adam optimizer along with gradient descent to achieve high-precision end-to-end joint angle estimation.

\begin{figure}[ht]
    \centering
    \includegraphics[width=0.75\linewidth]{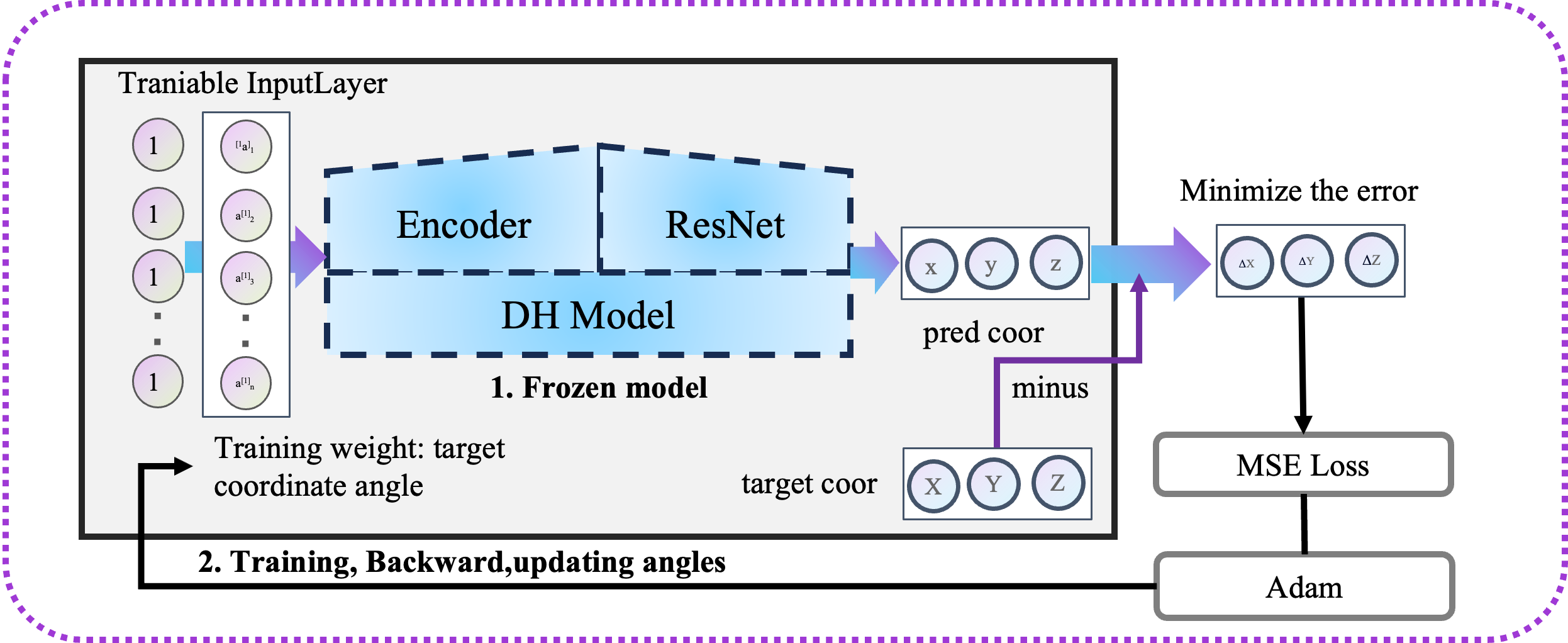}
    \caption{Inverse Angle Solving Algorithm Pipeline}
    \label{fig:2.5}
\end{figure}

The overall algorithm pipeline is illustrated in Fig \ref{fig:2.5}.

\subsubsection{Algorithm Steps}
\label{sec:2.4.1}

The inverse joint compensation algorithm proceeds as follows (see Fig.~\ref{fig:2.5}):

\begin{enumerate}[\textbf{Step 1:}]
    \item \textbf{Load and Freeze the Pre-trained BoTER Model} \\[0.5ex]
    Load the BoTER model with the best-performing parameters, freeze all its weights to prevent updates, and switch the model to evaluation mode to ensure stable inference behavior.

    \item \textbf{Design a Trainable Input Layer} \\[0.5ex]
    Add a trainable 6-dimensional input layer before the frozen BoTER model. During the forward pass, the input vector is first processed by this layer and then passed through the BoTER model to produce a predicted end-effector position. The final output is the 3D error vector between the predicted and target positions.

    \item \textbf{Initialize Input Layer Parameters} \\[0.5ex]
    Use the theoretical target coordinates as the ground truth, and their corresponding six joint angles as the initial guess. Convert these angles into radians and define them as a trainable tensor with gradient tracking enabled. This tensor initializes the weights of the input layer.

    Use the Adam optimizer with a learning rate of $1 \times 10^{-4}$, and restrict parameter updates to the input layer only. Adam is selected due to its adaptive momentum properties, which improve convergence in non-convex optimization landscapes.

    \item \textbf{Iterative Optimization Process} \\[0.5ex]
    Set the maximum number of iterations to 500. In each iteration, perform the following steps:

    \begin{enumerate}[(1)]
        \item \textbf{Forward Pass} \\[0.5ex]
        Define the optimization target as a zero vector in $\mathbb{R}^3$. Feed a 6D all-ones tensor into the input layer, and compute the predicted end-effector position through the BoTER model. Calculate the mean squared error (MSE) between the predicted and target positions as the loss.

        \item \textbf{Backpropagation and Weight Update} \\[0.5ex]
        Compute the gradient of the loss with respect to the input layer weights. Perform a gradient descent step using Adam. To prevent gradient explosion, utilize Adam’s internal gradient clipping mechanism to implicitly constrain the update magnitude.

        \item \textbf{Early Stopping} \\[0.5ex]
        If the MSE loss drops below a threshold of $1 \times 10^{-4}$, terminate the iteration early to accelerate computation.
    \end{enumerate}
\end{enumerate}

\subsubsection{Post-Processing}
\label{sec:2.4.2}

After optimization, the weights of the input layer represent the final joint angles. These are converted from radians back to degrees, and angle compensation is computed by subtracting the initial joint angles.

\[
\Delta \theta = \theta_{\text{final-rad}} \times \frac{180}{\pi} - \theta_{\text{initial-deg}} \tag{19}
\]

The theoretical position is used as the target. When evaluated in the UR robotic arm dataset, the algorithm converges on average after 147 iterations.

\section{Experiments}
\label{sec:3}

To evaluate the performance of the SPI-BoTER model in predicting the end-effector position of industrial robotic arms, we built a high-precision data acquisition platform based on a 6-DOF UR5 robotic arm and the TrackScan Sharp series 3D scanning system. This chapter provides a brief overview of the dataset acquisition process, including the experimental setup and data collection procedure, as well as the training process of the SPI-BoTER model. The goal is to provide reliable data support for the training and validation of the error compensation algorithm.

\subsection{Data Collection and Calibration Process}
\label{sec:3.1}

This research utilized a six-axis UR5 robotic manipulator for conducting experimental measurements. Weighing about 18.4 kilograms, the UR5 demonstrates an adept capability to manage loads up to 5 kilograms. Under standard operating conditions, it consumes roughly 150 watts of power and boasts a positional repeatability accuracy of $±0.10$ millimeters, reflecting its precision capabilities. Data acquisition was facilitated through the use of the TrackScan Sharp series of three-dimensional tracking scanning systems alongside TViewer software. The scanning system integrates binocular vision technology with a handheld scanner and is equipped with a high-resolution industrial camera that features 25 megapixels. In addition, it incorporates an edge computing module for real-time data processing. The system achieves a high-precision measurement volume of up to 135 cubic meters, with a maximum scanning range extending to 8.5 meters. Detailed specifications are provided in Table \ref{tab:2}. 

\begin{table}[ht]
  \centering
  \begin{tabular}{ccc}
      \hline
      \multicolumn{2}{c}{Parameters} & Values\\
      \hline
      \multicolumn{2}{c}{Maximum scan rate} & 4,860,000 $times/s$\\
      \multicolumn{2}{c}{Maximum scan format} & 800 $mm$$\times$700 $mm$\\
      \multicolumn{2}{c}{Highest resolution} & 0.02 $mm$\\
      \multirow{4}{*}{Calculate volumetric accuracy} & 10.4$m^3$ (3.5 $m$) & 0.048 $mm$\\
      &35$m^3$ (5.3 $m$) & 0.069 $mm$\\
      &90$m^3$ (7.2 $m$) & 0.128 $mm$\\
      &135$m^3$(8.5 $m$) & 0.159 $mm$\\
      \multicolumn{2}{c}{Datum distance} & 300 $mm$\\
      \multicolumn{2}{c}{Depth of field} & 400 $mm$,800 $mm$\\
      \multicolumn{2}{c}{Operating temperature} & -10 $\sim$ $40^\circ C$\\
      \hline
  \end{tabular}
  \caption{Technical Specifications for the TrackScan Sharp Series of Tracking 3D Scanning Systems}
  \label{tab:2}
\end{table}
\begin{figure}[ht]
    \centering
    \includegraphics[width=0.75\linewidth]{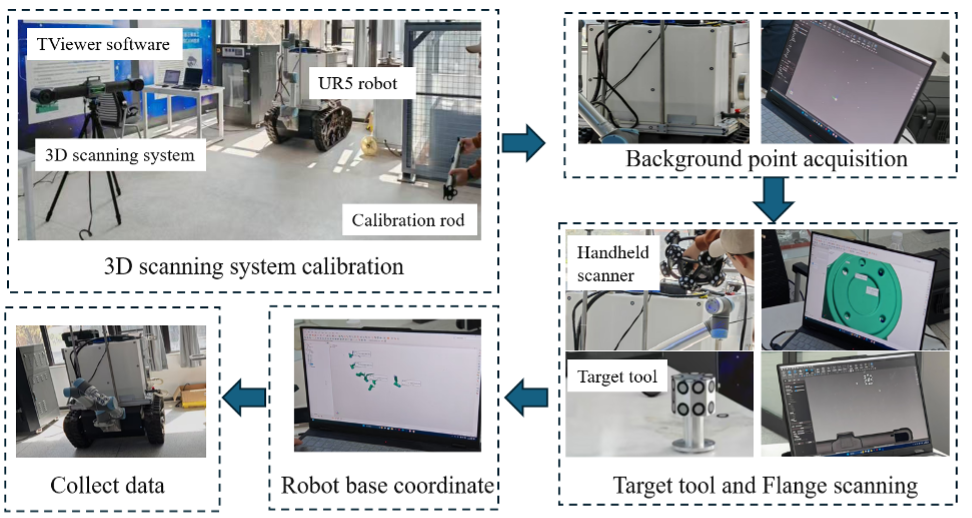}
    \caption{Experiments of error data collection}
    \label{fig:3.1}
\end{figure}

During the testing process, a mechanism equipped with visual markers was attached to the flange of the robot’s end-effector. The coordinates of the marker mechanism were determined using a tracking three-dimensional scanning system, as illustrated in Figure~\ref{fig:3.1}. Initially, calibration of both the three-dimensional scanning system and the robotic platform was performed. The detailed procedure is outlined below:

\begin{itemize}
    \item \textbf{Calibration of the Three-Dimensional Scanning System:} All equipment connections, including those between the 3D scanner, computer, and associated control software, were verified to ensure proper functionality. Calibration was conducted in a stable environment using a calibration rod to correct optical distortions and systemic biases, thereby minimizing absolute measurement errors.

    \item \textbf{Collection of Background Points:} Background point data within the working area were acquired using the 3D scanning system. This step aimed to capture information about static objects in the workspace, mitigating errors caused by environmental factors.

    \item \textbf{Scanning the Flange and Target Tool:} High-contrast markers were securely mounted on the robot's end effector (flange). Both the flange and the target tool were independently scanned to establish clear reference points for the 3D scanning process, further reducing measurement uncertainties.

    \item \textbf{Calibration of the Robot Base Coordinate System:} The base coordinate system within the robot’s workspace was defined, anchored to a fixed workbench. Precise alignment of this coordinate system minimized relative errors arising from scaling discrepancies or other inconsistencies.
\end{itemize}

Let the coordinates of a point in the base coordinate system (with the base as the origin) be $Q_i = (X_i, Y_i, Z_i)$, and its corresponding coordinates in the world coordinate system be $P_i = (x_i, y_i, z_i)$. The transformation relationship from the base coordinate system to the world coordinate system is given by:
\begin{equation}
    P_i = R \cdot Q_i + T
    \label{eq:transformation}\tag{20}
\end{equation}
where $R$ is the rotation matrix and $T$ is the translation vector.

To eliminate the influence of translation, the centroids of the base and world coordinate systems are computed:
\begin{equation}
    \bar{Q} = \frac{1}{n} \sum_{i=1}^{n} Q_i, \quad
    \bar{P} = \frac{1}{n} \sum_{i=1}^{n} P_i
    \label{eq:centroids}\tag{21}
\end{equation}

The centered coordinates are obtained by:
\begin{equation}
    Q_i' = Q_i - \bar{Q}, \quad
    P_i' = P_i - \bar{P}
    \label{eq:centered_coords}\tag{22}
\end{equation}

The covariance matrix $H$ is constructed as:
\begin{equation}
    H = \frac{1}{n} \sum_{i=1}^{n} (Q_i')^{T} \cdot P_i'
    \label{eq:covariance}\tag{23}
\end{equation}

Singular value decomposition (SVD) is applied to matrix $H$:
\begin{equation}
    H = U \cdot \Sigma \cdot V^T
    \label{eq:svd}\tag{24}
\end{equation}
where $U$ and $V$ are orthogonal matrices and $\Sigma$ is a diagonal matrix containing the singular values.

The rotation matrix $R$ is then constructed as:
\begin{equation}
    R = V 
    \begin{bmatrix}
        I_{2 \times 2} & 0 \\
        0 & \det(VU^T)
    \end{bmatrix}
    U^T
    \label{eq:rotation}\tag{25}
\end{equation}
to ensure a right-handed coordinate system (i.e., $\det(R)=1$).

Finally, the translation vector $T$ is computed by:
\begin{equation}
    T = \bar{P} - R \cdot \bar{Q}
    \label{eq:translation}\tag{26}
\end{equation}

During the data acquisition process, the 3D scanning system was placed 3 meters away from the robot. The robot was programmed to move to a specific location and pause for 10 seconds, during which the 3D scanning system recorded the coordinates of the flange center. The robot then proceeded to the next location and this process was repeated until a total of 800 positions were measured. Among these, 724 sets were selected as the calibration dataset.

These points were randomly distributed across the robot's frontal workspace. For each position, the joint angles of the robot were recorded and used in the forward kinematic model to calculate the theoretical positions. The actual positions, relative to the coordinate frame of the scanning system, were obtained directly using the measurement device and recorded using the TViewer software.

\subsection{Model Training Process}
\label{sec:3.2}

\begin{figure}[ht]
    \centering
    \includegraphics[width=0.5\textwidth]{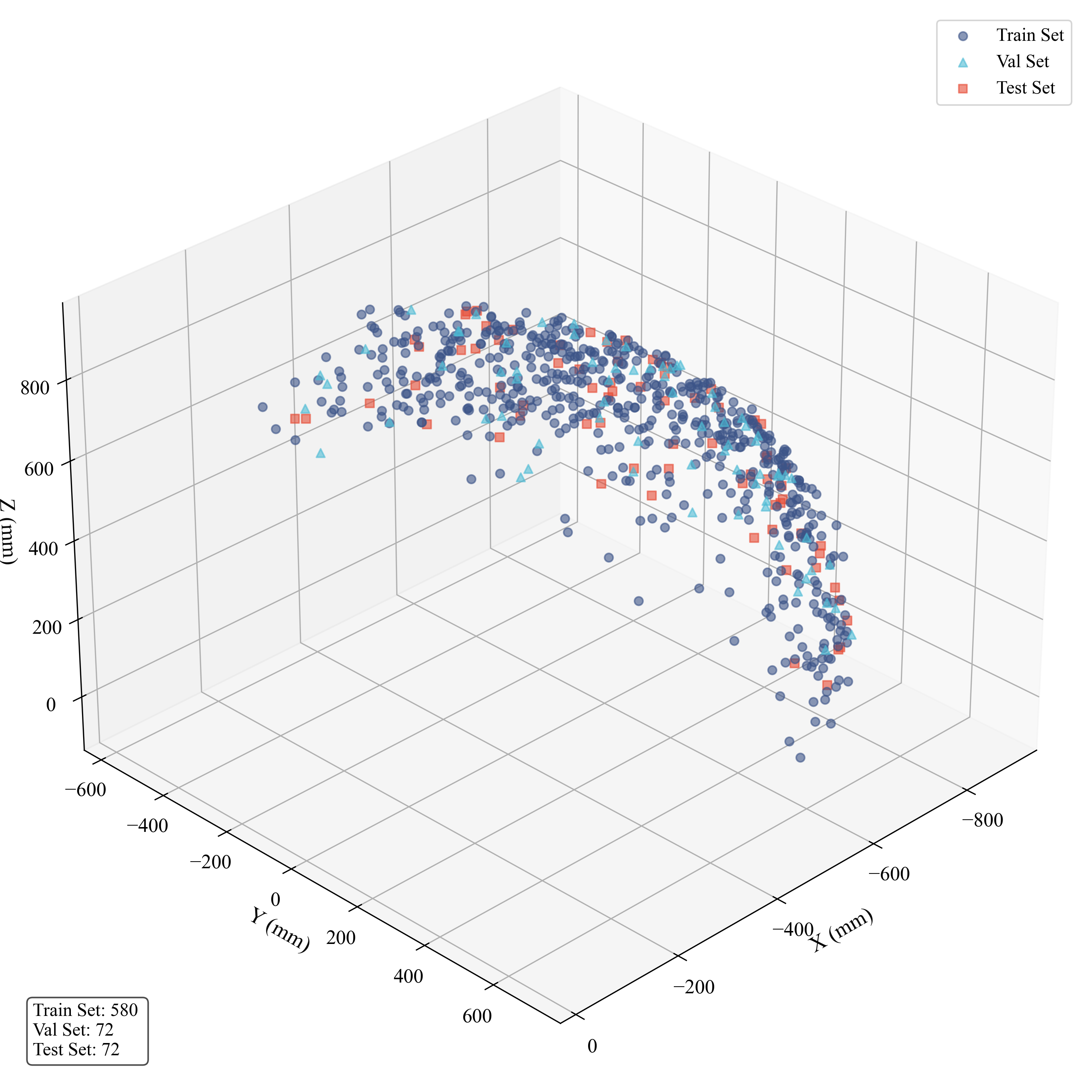}
    \caption{3D sample distribution}
    \label{fig:3.2}
\end{figure}

The SPI-BoTER model training was carried out on a workstation equipped with Ubuntu 22.04.5 LTS, a NVIDIA RTX A6000 GPU, and an Intel(R) Xeon(R) Platinum 8352V CPU. The dataset collected by the industrial robot was randomly split into training, validation, and testing sets in a ratio of 8:1:1, using a random seed of 7 (see Fig.~\ref{fig:3.2}).

The detailed training procedure is as follows:

\begin{enumerate}[(1)]
    \item \textbf{Initialization:} The BoTER model weights were initialized using a random seed of 139. Based on the kinematic structure of the UR5 robotic arm, a sparse self-attention mask was designed accordingly.

    \item \textbf{Loss Function Definition:} An SPI loss function was constructed for the model, which integrates both data residuals and spatial physical residuals. This ensures that the model not only accurately predicts the end-effector positions, but also adheres to physical constraints.

    \item \textbf{Backpropagation and Weight Update:} The Adam optimizer was applied to compute the gradients of the loss function through backpropagation. These gradients were then used to update the network parameters, ensuring that the model progressively minimizes loss during each training epoch.

    \item \textbf{Iterative Training and Model Checkpointing:} The model was trained iteratively, with performance dynamically evaluated during training. The optimal model weights were periodically saved for future testing and deployment.
\end{enumerate}

After 150 epochs of training, the model achieved its best performance in epoch 111. The optimal hyperparameter configuration of the SPI-BoTER model is summarized as follows:
\begin{figure}[ht]
    \centering
    \includegraphics[width=0.5\textwidth]{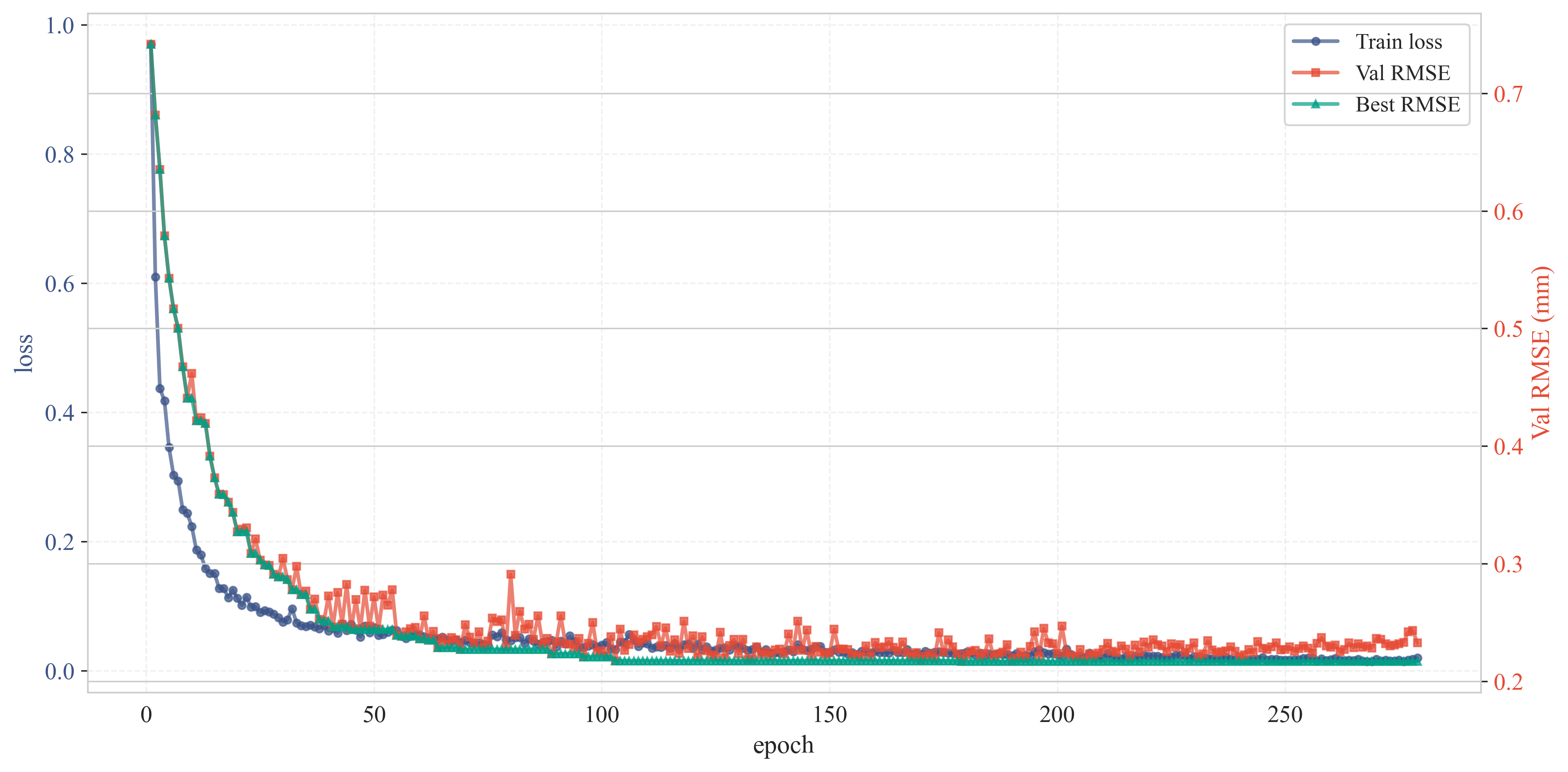}
    \caption{Training process of SPI-BoTER}
    \label{fig:3.3}
\end{figure}
\begin{itemize}
    \item \textbf{Input/Output Dimensions:} 6-dimensional joint angle input and 3-dimensional end-effector position output.
    
    \item \textbf{Transformer Encoder:} Input dimension $d_{\text{model}} = 126$, number of encoder layers $N_{\text{layer}} = 4$, number of attention heads $n_{\text{head}} = 9$.
    
    \item \textbf{Residual Network:} Hidden layer dimension $d_{\text{hidden}} = 512$.
    
    \item \textbf{Training Configuration:} The optimizer used was AdamW, with a learning rate of $1 \times 10^{-4}$, weight decay of $1 \times 10^{-6}$, batch size of 256, and a maximum of 5000 training epochs. Gradient clipping was applied with a threshold of 1.0.
\end{itemize}

\section{Results and Discussion}
\subsection{Performance Metrics}
\label{sec:4.1}

To quantitatively evaluate the performance of different models, the individual loss components in SPI, and various structural designs in BoTER on the task of predicting the end-effector coordinates of a six-axis industrial robot, we adopt four metrics: Mean Absolute Error (MAE), Coefficient of Determination ($R^2$), Mean Squared Error (MSE), and Root Mean Squared Error (RMSE)~\cite{42}. The corresponding formulations are as follows:

\begin{equation}
    \text{MAE}_x = \frac{1}{n} \sum_{i=1}^n \left| \hat{x}_i - x_{\text{real},i} \right| \tag{27}
\end{equation}
\begin{equation}
    \text{MAE}_y = \frac{1}{n} \sum_{i=1}^n \left| \hat{y}_i - y_{\text{real},i} \right| \tag{28}
\end{equation}
\begin{equation}
    \text{MAE}_z = \frac{1}{n} \sum_{i=1}^n \left| \hat{z}_i - z_{\text{real},i} \right| \tag{29}
\end{equation}
\begin{equation}
    \text{MAE}_{3d} = \frac{1}{n} \sum_{i=1}^n \sqrt{(\hat{x}_i - x_{\text{real},i})^2 + (\hat{y}_i - y_{\text{real},i})^2 + (\hat{z}_i - z_{\text{real},i})^2} \tag{30}
\end{equation}

\begin{equation}
    \text{MSE}_x = \frac{1}{n} \sum_{i=1}^n (\hat{x}_i - x_{\text{real},i})^2 \tag{31}
\end{equation}
\begin{equation}
    \text{MSE}_y = \frac{1}{n} \sum_{i=1}^n (\hat{y}_i - y_{\text{real},i})^2 \tag{32}
\end{equation}
\begin{equation}
    \text{MSE}_z = \frac{1}{n} \sum_{i=1}^n (\hat{z}_i - z_{\text{real},i})^2 \tag{33}
\end{equation}
\begin{equation}
    \text{MSE}_{3d} = \frac{1}{n} \sum_{i=1}^n \left[(\hat{x}_i - x_{\text{real},i})^2 + (\hat{y}_i - y_{\text{real},i})^2 + (\hat{z}_i - z_{\text{real},i})^2\right] \tag{34}
\end{equation}

\begin{equation}
    \text{RMSE}_x = \sqrt{\frac{1}{n} \sum_{i=1}^n (\hat{x}_i - x_{\text{real},i})^2} \tag{35}
\end{equation}
\begin{equation}
    \text{RMSE}_y = \sqrt{\frac{1}{n} \sum_{i=1}^n (\hat{y}_i - y_{\text{real},i})^2} \tag{36}
\end{equation}
\begin{equation}
    \text{RMSE}_z = \sqrt{\frac{1}{n} \sum_{i=1}^n (\hat{z}_i - z_{\text{real},i})^2} \tag{37}
\end{equation}
\begin{equation}
    \text{RMSE}_{3d} = \sqrt{\frac{1}{n} \sum_{i=1}^n \left[(\hat{x}_i - x_{\text{real},i})^2 + (\hat{y}_i - y_{\text{real},i})^2 + (\hat{z}_i - z_{\text{real},i})^2\right]} \tag{38}
\end{equation}

\begin{equation}
    R^2_x = 1 - \frac{\sum_{i=1}^n \left( \widehat{\Delta x}_i - \Delta x_{\text{real},i} \right)^2}{\sum_{i=1}^n \left( \overline{\Delta x}_r - \Delta x_{\text{real},i} \right)^2} \tag{39}
\end{equation}
\begin{equation}
    R^2_y = 1 - \frac{\sum_{i=1}^n \left( \widehat{\Delta y}_i - \Delta y_{\text{real},i} \right)^2}{\sum_{i=1}^n \left( \overline{\Delta y}_r - \Delta y_{\text{real},i} \right)^2} \tag{40}
\end{equation}
\begin{equation}
    R^2_z = 1 - \frac{\sum_{i=1}^n \left( \widehat{\Delta z}_i - \Delta z_{\text{real},i} \right)^2}{\sum_{i=1}^n \left( \overline{\Delta z}_r - \Delta z_{\text{real},i} \right)^2} \tag{41}
\end{equation}
\begin{equation}
    R^2_{3d} = 1 - \frac{\sum_{i=1}^n \left[ (\widehat{\Delta x}_i - \Delta x_{\text{real},i})^2 + (\widehat{\Delta y}_i - \Delta y_{\text{real},i})^2 + (\widehat{\Delta z}_i - \Delta z_{\text{real},i})^2 \right]}{\sum_{i=1}^n \left[ (\overline{\Delta x}_r - \Delta x_{\text{real},i})^2 + (\overline{\Delta y}_r - \Delta y_{\text{real},i})^2 + (\overline{\Delta z}_r - \Delta z_{\text{real},i})^2 \right]} \tag{42}
\end{equation}

Here, $n$ denotes the number of samples; $x_{\text{real},i}$, $y_{\text{real},i}$, and $z_{\text{real},i}$ are the ground-truth XYZ coordinates of the $i$-th sample, and $\hat{x}_i$, $\hat{y}_i$, and $\hat{z}_i$ are the corresponding predicted values. $\overline{\Delta x}_r$, $\overline{\Delta y}_r$, and $\overline{\Delta z}_r$ represent the arithmetic mean of the absolute prediction errors for each dimension, and are used in the computation of the coefficient of determination $R^2$.

Mean Squared Error (MSE) measures the average squared difference between predicted and actual values; the smaller the value, the better the model's performance. Root Mean Squared Error (RMSE), as the square root of MSE, provides an error metric with the same dimensional unit as the original data, making the interpretation more intuitive. The coefficient of determination ($R^2$) evaluates how well the model explains the variability of the data, ranging from 0 to 1, with higher values indicating better model fit. Mean Absolute Error (MAE) calculates the average absolute deviation between predictions and actual values, offering an easily interpretable measure of predictive accuracy. Together, these metrics provide a comprehensive evaluation of model performance, allowing a full understanding of both its predictive ability and its adaptation capacity~\cite{42}.

\subsection{Performance on the UR5 Industrial Robot Random Point Dataset}
\label{sec:4.2}

We designed four distinct sets of experiments to evaluate the accuracy and superiority of the SPI-BoTER model, as well as two ablation studies to assess the effectiveness of individual loss components in SPI and the module design choices in BoTER. All experiments were conducted on a workstation running Ubuntu 22.04.5 LTS, equipped with an NVIDIA RTX A6000 GPU and an Intel(R) Xeon(R) Platinum 8352V CPU. The dataset, collected using a UR series industrial robot, contains a total of 724 samples. The dataset was randomly split into training, validation, and test sets with a ratio of 8:1:1 using a random seed of 7.

\subsubsection{Accuracy of the SPI-BoTER Model in Predicting End-Effector Coordinates}
\label{sec:4.2.1}

The aim of this experiment is to evaluate the precision and accuracy of the SPI-BoTER model in predicting the end-effector coordinates of an industrial robot. The experimental results demonstrate the model's performance across the training, validation, and test sets. The prediction performance of the SPI-BoTER model is summarized in Table~\ref{tab:BoTER_results}.

\vspace{1em}
\begin{figure}[ht]
    \centering
    \includegraphics[width=0.5\textwidth]{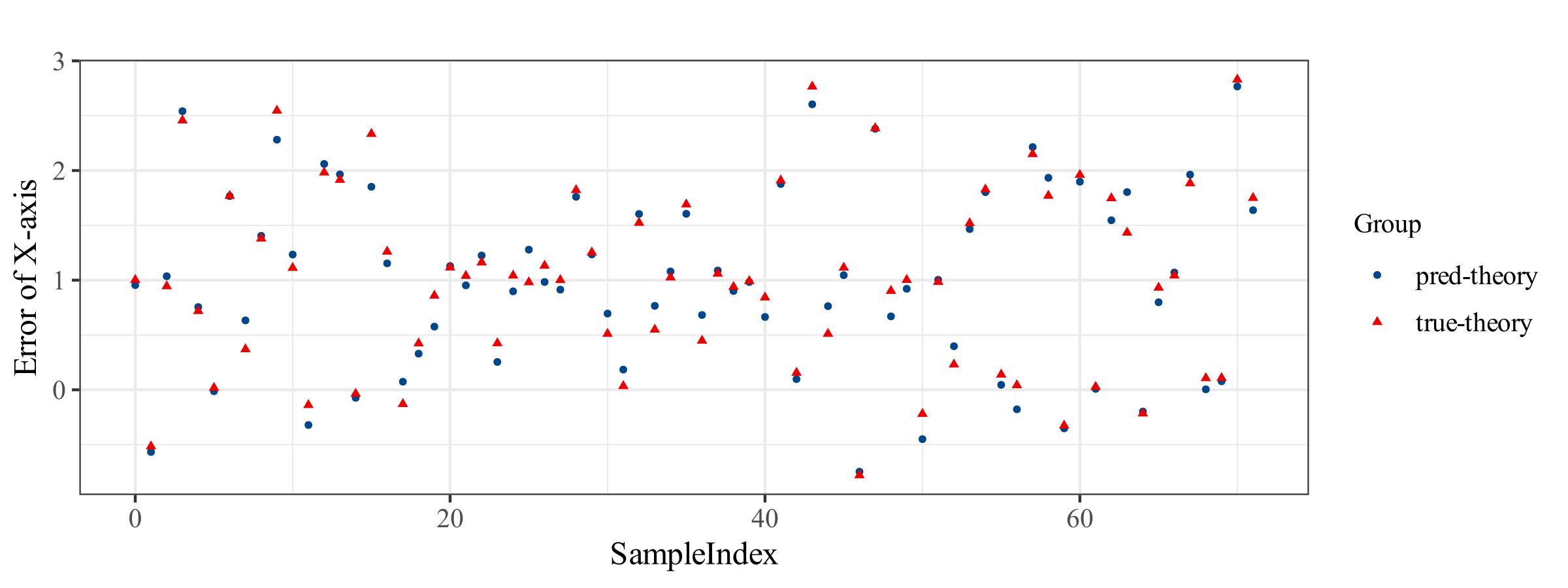}
    \caption{True vs. Predicted Error Scatter Chart on the X-axis (Test Set)}
    \label{fig:scatter_x}
\end{figure}

\begin{figure}[ht]
    \centering
    \includegraphics[width=0.5\textwidth]{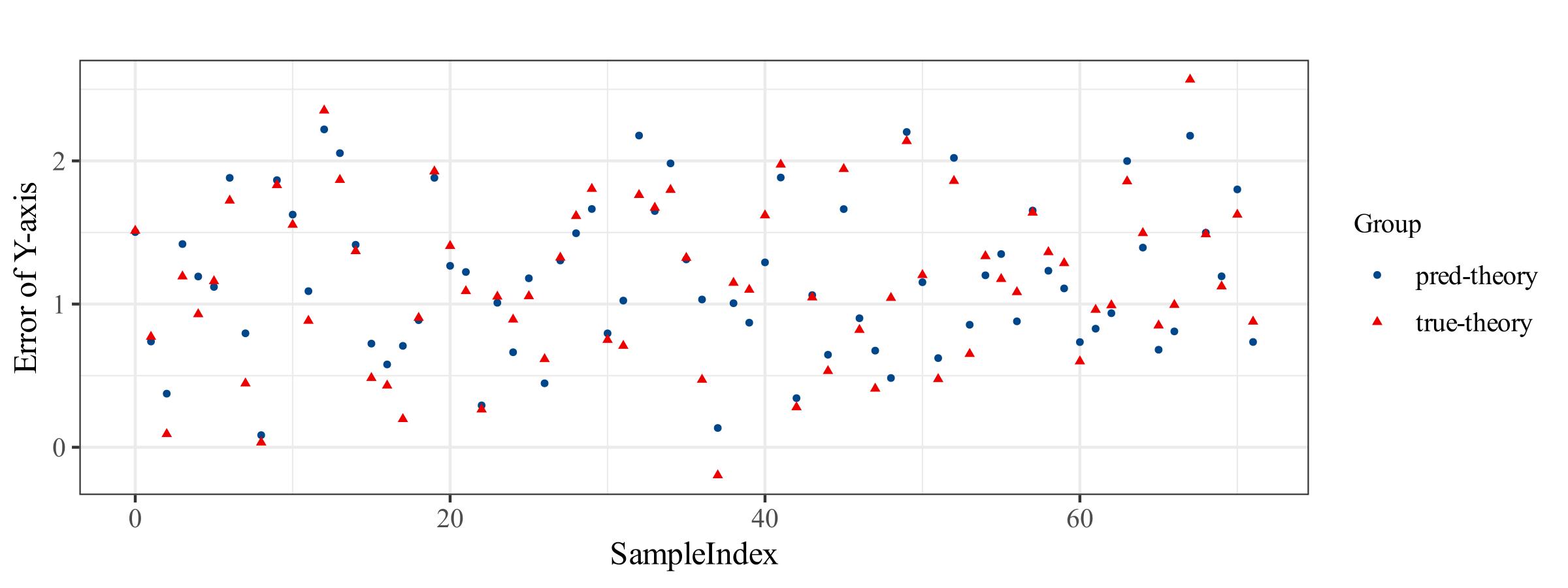}
    \caption{True vs. Predicted Error Scatter Chart on the Y-axis (Test Set)}
    \label{fig:scatter_y}
\end{figure}

\begin{figure}[ht]
    \centering
    \includegraphics[width=0.5\textwidth]{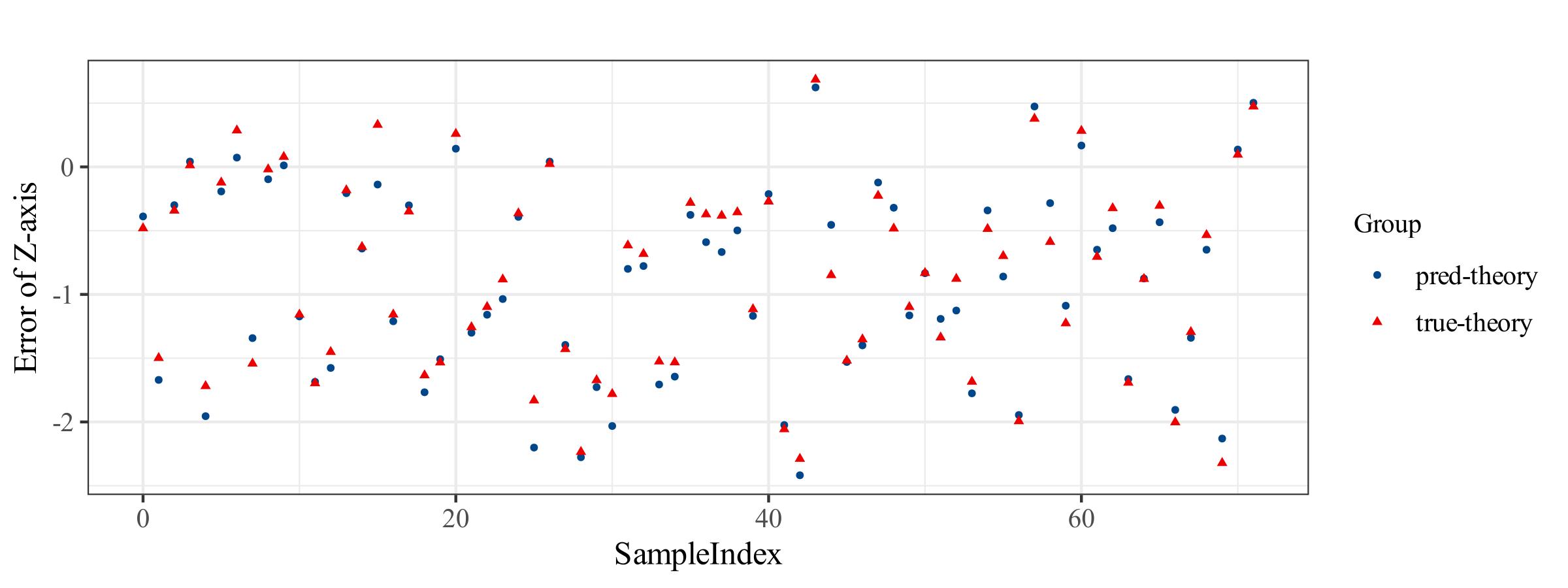}
    \caption{True vs. Predicted Error Scatter Chart on the Z-axis (Test Set)}
    \label{fig:scatter_z}
\end{figure}

\begin{figure}[h!]
    \centering
    \includegraphics[width=0.5\textwidth]{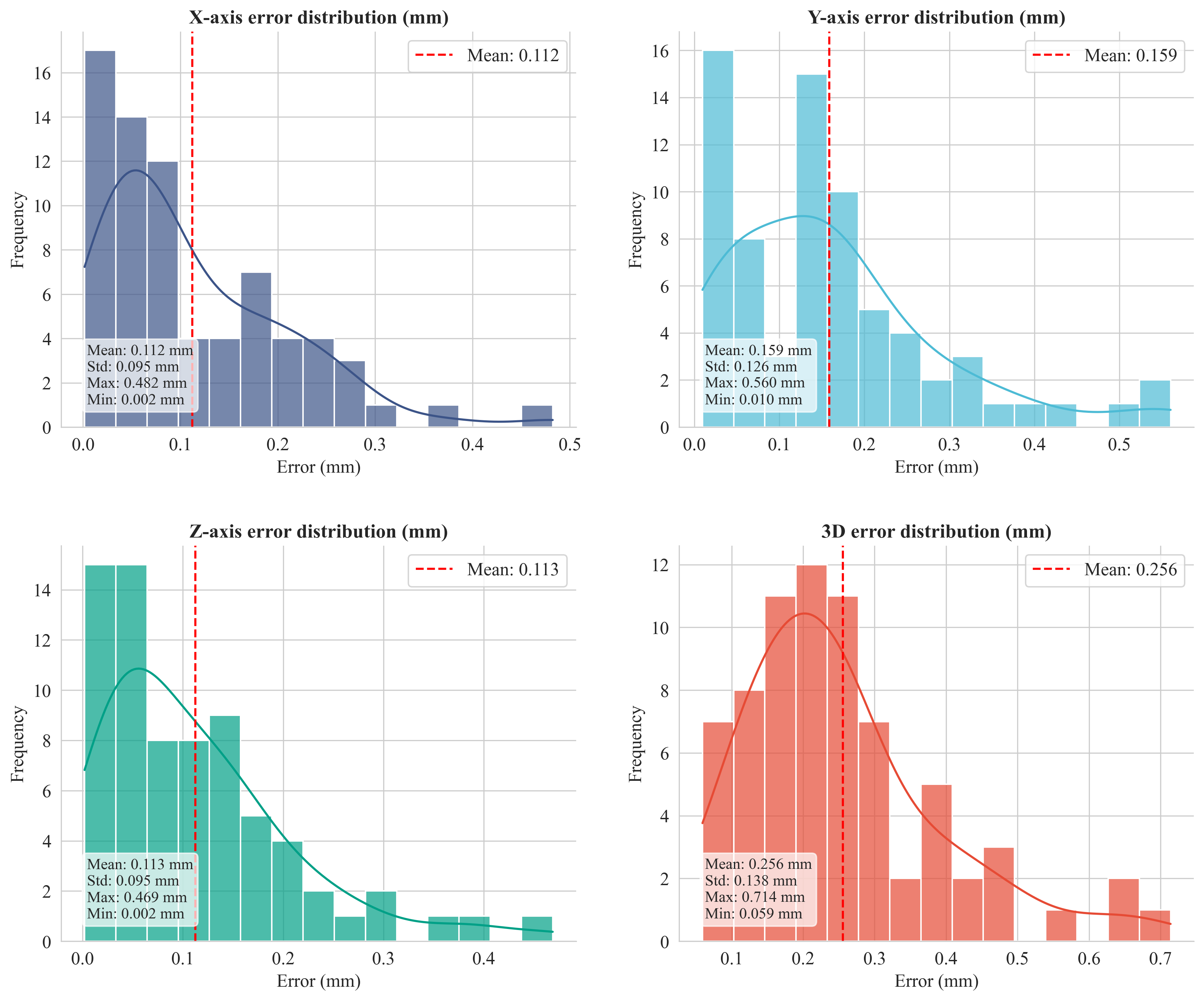}
    \caption{Error Distribution on the Test Set}
    \label{fig:error_distribution}
\end{figure}

\begin{table}[h!]
\centering
\caption{Performance Metrics of the SPI-BoTER Model}
\begin{tabular}{ccccccc}
\hline
&Samples &Axis & MAE(mm) & MSE(mm$^2$) & RMSE(mm) & $R^2$ \\
\hline
\multirow{4}{*}{Train} & \multirow{4}{*}{580}
& X & 0.0190 & 0.0190 & 0.1377 & 0.9719 \\
& & Y & 0.0422 & 0.0422 & 0.2054 & 0.8967 \\
& & Z & 0.0175 & 0.0175 & 0.1325 & 0.9726 \\
& & 3D & 0.2189 & 0.0787 & 0.2805 & 0.9543 \\
\hline
\multirow{4}{*}{Train} & \multirow{4}{*}{580}
& X & 0.1580 & 0.0432 & 0.2079 & 0.9342 \\
& & Y & 0.1905 & 0.0656 & 0.2562 & 0.8236 \\
& & Z & 0.1444 & 0.0321 & 0.1792 & 0.9351 \\
& & 3D & 0.3218 & 0.1410 & 0.3755 & 0.9075 \\
\hline
\multirow{4}{*}{Train} & \multirow{4}{*}{580}
& X & 0.1110 & 0.0213 & 0.1461 & 0.9682 \\
& & Y & 0.1537 & 0.0379 & 0.1947 & 0.8822 \\
& & Z & 0.1135 & 0.0220 & 0.2170 & 0.9612 \\
& & 3D & 0.2515 & 0.2850 & 0.0812 & 0.9479 \\
\hline
\end{tabular}
\label{tab:BoTER_results}
\end{table}

Figures~\ref{fig:scatter_x}–\ref{fig:scatter_z} illustrate the scatter plots of predicted versus actual errors for the X, Y and Z axes of the test set. These plots compare the differences between the predicted and theoretical coordinates with those between the measured and theoretical coordinates in the error space, demonstrating that the predicted values closely approximate the measured values. Figure~\ref{fig:error_distribution} presents the error distribution in the test set. The performance metrics summarized in Table~\ref{tab:performance} show that the SPI-BoTER model achieves low values for MSE, MAE and RMSE in training, validation and test sets. Except for the Y-axis, the $R^2$ values exceed 0.96, indicating that the model exhibits high accuracy in predicting the coordinates of the end effector of the robotic arm.

\subsubsection{Performance Comparison between DNN and SPI-BoTER}
\label{sec:4.2.2}

This experiment aims to evaluate the effectiveness of the SPI-BoTER framework on the task of predicting the 3D coordinates of an industrial robot end effector. We compared our approach with two baseline methods: the standard Deep Neural Network (DNN) \cite{43} and GPSO-DNN \cite{46}. All models were implemented and reproduced on the same random-point dataset of UR robot, consisting of 724 samples. The parameter configurations for both DNN and GPSO-DNN strictly follow those reported in their original publications.

\begin{figure}[h!]
    \centering
    \includegraphics[width=0.6\textwidth]{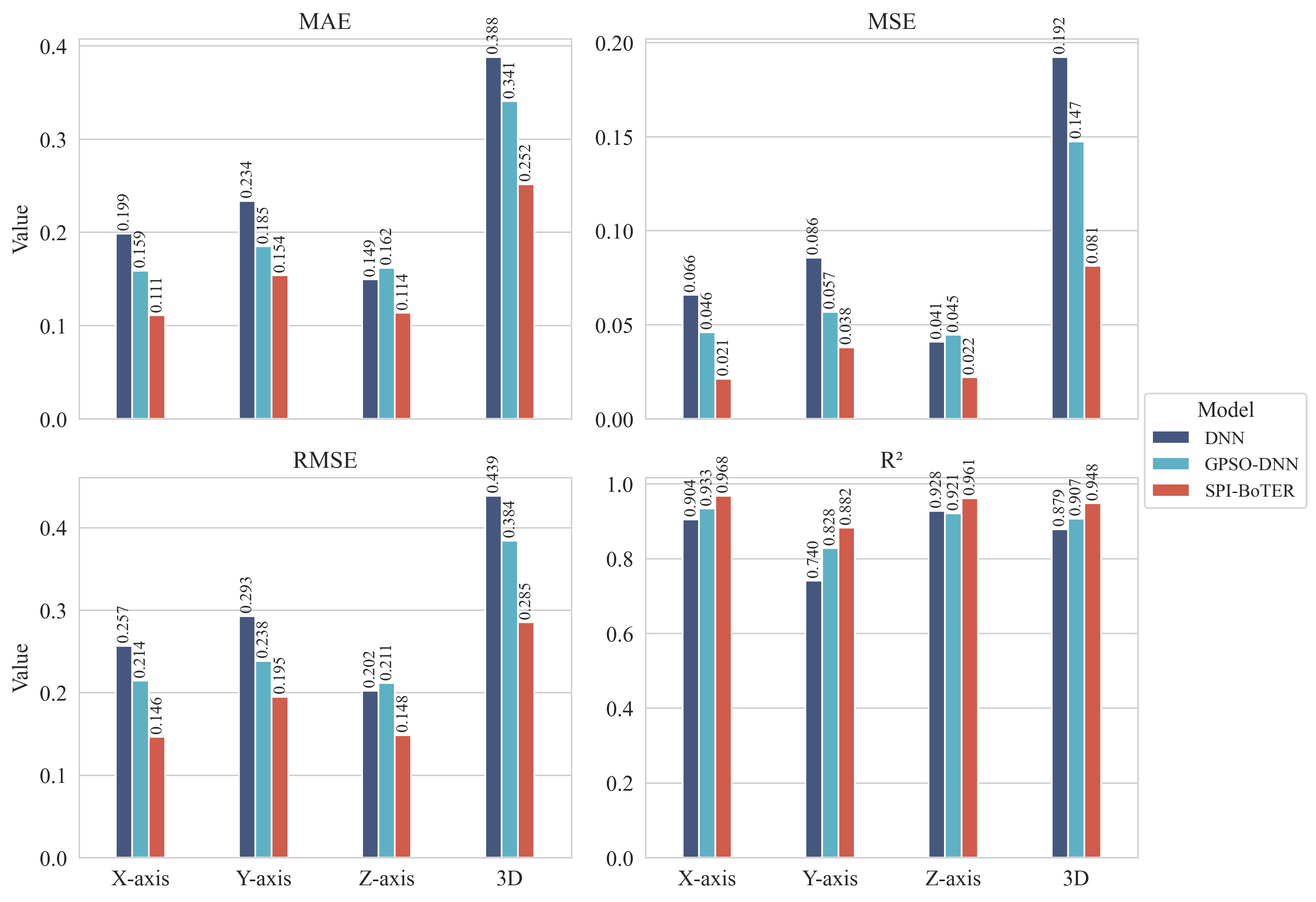}
    \caption{Performance Comparison between DNN and SPI-BoTER across MAE, MSE, RMSE, and $R^2$}
    \label{fig:compare_models}
\end{figure}

All models were trained under the same hardware environment (NVIDIA RTX A6000 GPU) and consistent training settings, using Adam Optimizer with a learning rate of $1 \times 10^{-4}$ and a batch size of 256. The comparison of error metrics on the test set is summarized in Table~\ref{tab:model_comparison}. Figure~\ref{fig:compare_models} visualizes the performance of the models in the MAE, MSE, RMSE, and $R^2$ metrics.

\vspace{1em}

\begin{table}[h!]
\centering
\caption{Test Set Error Comparison among DNN, GPSO-DNN, and SPI-BoTER}
\label{tab:model_comparison}
\begin{tabular}{cccccc}
\hline
{Method} & {Axis} & {MAE (mm)} & {MSE (mm\textsuperscript{2})} & {RMSE (mm)} & {$R^2$} \\
\hline
\multirow{4}{*}{DNN} 
& X & 0.1986 & 0.0658 & 0.2566 & 0.9045 \\
& Y & 0.2336 & 0.0856 & 0.2926 & 0.7404 \\
& Z & 0.1493 & 0.0409 & 0.2022 & 0.9278 \\
& 3D & 0.3880 & 0.1923 & 0.4386 & 0.8787 \\
\hline
\multirow{4}{*}{GPSO-DNN} 
& X & 0.1588 & 0.0460 & 0.2144 & 0.9333 \\
& Y & 0.1848 & 0.0568 & 0.2384 & 0.8277 \\
& Z & 0.1618 & 0.0446 & 0.2113 & 0.9211 \\
& 3D & 0.3405 & 0.1474 & 0.3840 & 0.9070 \\
\hline
\multirow{4}{*}{SPI-BoTER} 
& X & 0.1110 & 0.0213 & 0.1461 & 0.9682 \\
& Y & 0.1537 & 0.0379 & 0.1947 & 0.8822 \\
& Z & 0.1135 & 0.0220 & 0.1484 & 0.9612 \\
& 3D & 0.2515 & 0.0812 & 0.2850 & 0.9479 \\
\hline
\end{tabular}
\end{table}

For the DNN method, the mean absolute errors (MAEs) along the X, Y, and Z axes, and in the 3D space, are 0.1986 mm, 0.2336 mm, 0.1493 mm, and 0.3880 mm, respectively. Although GPSO-DNN improves on DNN with MAEs of 0.1588 mm (X), 0.1848 mm (Y), 0.1618 mm (Z) and 0.3405 mm (3D), its performance still lags behind SPI-BoTER, which achieves 0.1110 mm, 0.1537 mm, 0.1135 mm and 0.2515 mm for the corresponding dimensions. These results indicate that both DNN and GPSO-DNN exhibit greater variability and larger prediction errors in 3D coordinate estimation compared to SPI-BoTER.

SPI-BoTER outperforms the other models in all four evaluation metrics. Its 3D MAE of 0.2515 mm represents a reduction of 35.16\% compared to DNN and 26.14\% compared to GPSO-DNN. Similarly, its 3D $R^2$ score of 0.9479 shows an improvement of 7.87\% and 4.51\% over DNN and GPSO-DNN, respectively. The low MAE values across all axes (X: 0.1110 mm, Y: 0.1537 mm, Z: 0.1135 mm) further validate the stability of the SPI-BoTER predictions, demonstrating its effectiveness in suppressing anomalies in the 3D coordinate space.

These experimental results indicate that the SPI-BoTER model, through its dual-branch architecture combining physical mechanisms and data-driven learning, along with a hybrid loss function, significantly enhances physical interpretability and robustness under extreme conditions. While maintaining the flexibility of DNN-based models, SPI-BoTER provides a more effective solution for compensating robotic arm positioning errors in small-sample scenarios.

\subsubsection{Ablation Study on Loss Terms in SPI}
\label{sec:4.2.3}

To validate the necessity of each loss term in the SPI, we performed an ablation study based on the BoTER model. The experimental group design is summarized in Table~\ref{tab:ablation_groups}.

\begin{table}[h!]
\centering
\caption{Design of Ablation Groups for SPI Loss Terms}
\label{tab:ablation_groups}
\begin{tabular}{lll}
\hline
\textbf{Group Name} & \textbf{Modification} & \textbf{Objective} \\
\hline
Baseline & All loss terms and adaptive parameters retained & Establish the performance benchmark \\
Data Only & Only retains $\mathcal{L}_{\text{data}}$, spatial physics loss set to 0 & Test the necessity of physical constraints \\
\hline
\end{tabular}
\end{table}

\begin{figure}[h!]
    \centering
    \includegraphics[width=0.6\textwidth]{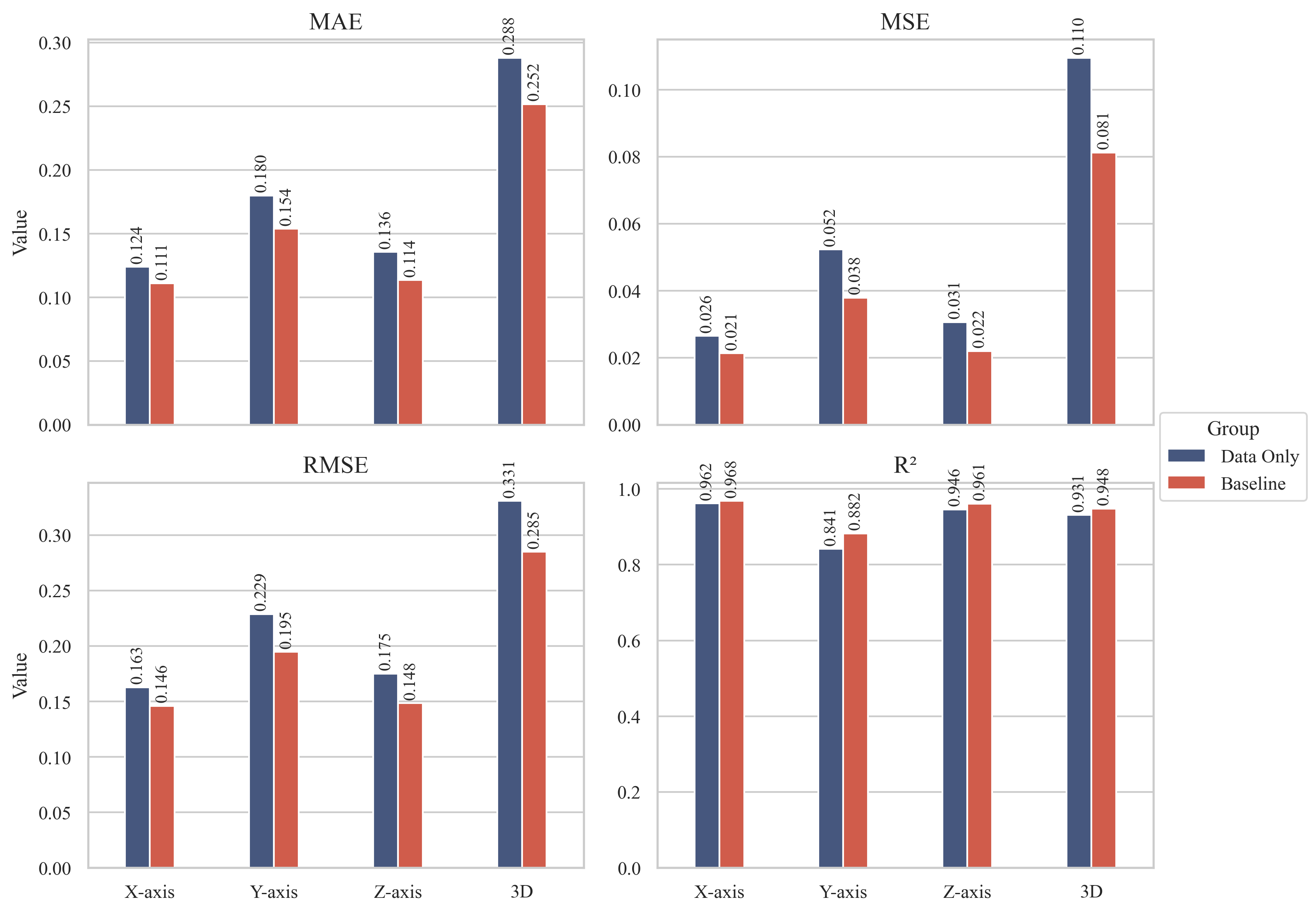}
    \caption{Performance Comparison of Ablation Groups for Each Loss Term of SPI}
    \label{fig:ablation_compare}
\end{figure}

As shown in Table~\ref{tab:ablation_groups}, the \textbf{ baseline} group retains both the spatial physics loss term and the adaptive parameters to establish a benchmark for performance comparison. In contrast, the \textbf{Data Only} group retains only the residual loss of data $\mathcal{L}_{\text{data}}$ while setting the loss based on physics to zero, in order to assess the importance of physical constraints. The experimental results are illustrated in Figure~\ref{fig:ablation_compare} and detailed in Table~\ref{tab:ablation_results}.

\vspace{1em}

\begin{table}[h!]
\centering
\caption{Performance Comparison of SPI Loss Term Ablation Groups}
\label{tab:ablation_results}
\begin{tabular}{cccccc}
\hline
{Group} & {Axis} & {MAE (mm)} & {MSE (mm\textsuperscript{2})} & {RMSE (mm)} & {$R^2$} \\
\hline
\multirow{4}{*}{Data Only} 
& X & 0.1239 & 0.0265 & 0.1628 & 0.9615 \\
& Y & 0.1799 & 0.0523 & 0.2288 & 0.8414 \\
& Z & 0.1355 & 0.0306 & 0.1751 & 0.9459 \\
& 3D & 0.2880 & 0.1095 & 0.3309 & 0.9309 \\
\hline
\multirow{4}{*}{Baseline} 
& X & 0.1110 & 0.0213 & 0.1461 & 0.9682 \\
& Y & 0.1537 & 0.0379 & 0.1947 & 0.8822 \\
& Z & 0.1135 & 0.0220 & 0.1484 & 0.9612 \\
& 3D & 0.2515 & 0.0812 & 0.2850 & 0.9479 \\
\hline
\end{tabular}
\end{table}

From Table~\ref{tab:ablation_results}, it is evident that the baseline group, which includes the spatially physics-informed loss term, achieves a 3D mean absolute error (MAE) of 0.2515 mm. This represents a 12\% reduction compared to the 0.2880 mm of the Data Only group, which omits physical loss. These results demonstrate that incorporating the spatial physics-informed loss effectively suppresses anomalous errors in 3D space and enhances the overall performance of the BoTER model.

\subsubsection{Ablation Study on the Architecture of BoTER Model}
\label{sec:4.2.4}

To verify the necessity of each architectural component in the BoTER model, we conducted an ablation study using a unified SPI loss and a consistent training configuration with backpropagation optimization. The design of the experimental groups is presented in Table~\ref{tab:structure_ablation_groups}.

\begin{table}[h!]
\centering
\caption{Design of BoTER Model Structure Ablation Experiment Groups}
\label{tab:structure_ablation_groups}
\begin{tabular}{lp{6cm}p{6cm}}
\hline
\textbf{Group Name} & \textbf{Modification} & \textbf{Objective} \\
\hline
Baseline & All layers retained & Establish performance benchmark \\

No Mask & Remove the sparse self-attention mask & Evaluate necessity of the masking mechanism \\

Body Mask & Replace with BoT mask design & Evaluate superiority of the proposed masking strategy \\

No ResNet & Replace residual blocks in the prediction head with linear layers & Evaluate necessity of residual blocks in prediction head \\

No Transformer & Remove the Transformer encoder layer and use only ResNet head & Evaluate necessity of the Transformer encoder \\
\hline
\end{tabular}
\end{table}

As shown in Table~\ref{tab:structure_ablation_groups}, the \textbf{ baseline} group retains all modules in the BoTER model, including the sparse self-attention mask, the residual block-based prediction head, and the Transformer encoder, to serve as the performance benchmark. The \textbf{No Mask} group removes the sparse self-attention mask specifically designed for six-axis industrial robots, effectively using a standard Transformer encoder. The \textbf{Body Mask} group adopts the mask design principle from the Body Transformer~\cite{36}, defined as\ref{equ:43}:

\[
M_{\text{body}} =
\begin{bmatrix}
0 & 0 & 1 & 1 & 1 & 1 \\
0 & 0 & 0 & 1 & 1 & 1 \\
1 & 0 & 0 & 0 & 1 & 1 \\
1 & 1 & 0 & 0 & 0 & 1 \\
1 & 1 & 1 & 0 & 0 & 0 \\
1 & 1 & 1 & 1 & 0 & 0 \\
\end{bmatrix}
\tag{43}\label{equ:43}
\quad \text{where } 0 = \text{visible},\ 1 = \text{masked}
\]

\begin{figure}[h!]
    \centering
    \includegraphics[width=0.6\textwidth]{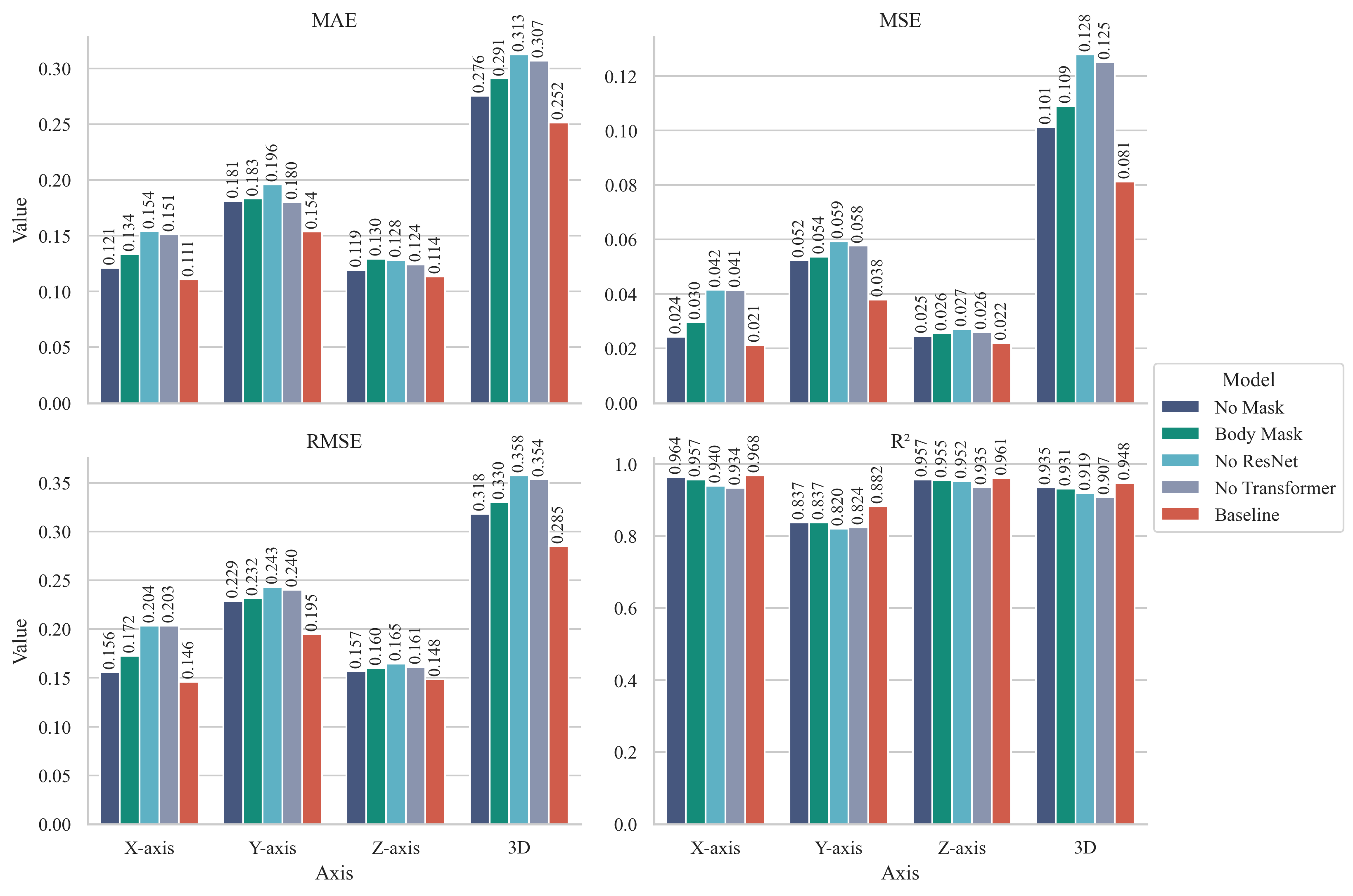}
    \caption{Performance Comparison of SPI-BoTER Model Structure Ablation Groups}
    \label{fig:structure_ablation_fig}
\end{figure}

The \textbf{No ResNet} group removes residual blocks in the prediction head and replaces them with fully connected linear layers, corresponding to a Transformer encoder with sparse masking but no residual prediction refinement. The \textbf{No Transformer} group completely eliminates the Transformer encoder and uses only ResNet to predict the end-effector coordinates. The experimental results are shown in Figure~\ref{fig:structure_ablation_fig} and Table~\ref{tab:structure_ablation_results}.

\vspace{1em}

\begin{table}[h!]
\centering
\caption{Performance Comparison of SPI-BoTER Model Structure Ablation Groups}
\label{tab:structure_ablation_results}
\begin{tabular}{ccccccc}
\hline
{Group} & {Epochs} & {Axis} & {MAE (mm)} & {MSE (mm$^2$)} & {RMSE (mm)} & {$R^2$} \\
\hline
\multirow{4}{*}{No Mask} & \multirow{4}{*}{402} 
& X & 0.1212 & 0.0243 & 0.1558 & 0.9639 \\
& & Y & 0.1813 & 0.0524 & 0.2289 & 0.8373 \\
& & Z & 0.1194 & 0.0246 & 0.1568 & 0.9567 \\
& & 3D & 0.2756 & 0.1012 & 0.3182 & 0.9352 \\
\hline
\multirow{4}{*}{Body Mask} & \multirow{4}{*}{103} 
& X & 0.1336 & 0.0297 & 0.1725 & 0.9568 \\
& & Y & 0.1833 & 0.0537 & 0.2317 & 0.8373 \\
& & Z & 0.1296 & 0.0256 & 0.1601 & 0.9547 \\
& & 3D & 0.2911 & 0.1090 & 0.3302 & 0.9312 \\
\hline
\multirow{4}{*}{No ResNet} & \multirow{4}{*}{304} 
& X & 0.1541 & 0.0415 & 0.2037 & 0.9398 \\
& & Y & 0.1960 & 0.0593 & 0.2434 & 0.8204 \\
& & Z & 0.1284 & 0.0271 & 0.1646 & 0.9521 \\
& & 3D & 0.3128 & 0.1279 & 0.3576 & 0.9193 \\
\hline
\multirow{4}{*}{No Transformer} & \multirow{4}{*}{341} 
& X & 0.1513 & 0.0414 & 0.2034 & 0.9342 \\
& & Y & 0.1803 & 0.0577 & 0.2402 & 0.8236 \\
& & Z & 0.1243 & 0.0260 & 0.1611 & 0.9351 \\
& & 3D & 0.3072 & 0.1250 & 0.3536 & 0.9075 \\
\hline
\multirow{4}{*}{Baseline} & \multirow{4}{*}{111} 
& X & 0.1110 & 0.0213 & 0.1461 & 0.9682 \\
& & Y & 0.1537 & 0.0379 & 0.1947 & 0.8822 \\
& & Z & 0.1135 & 0.0220 & 0.1484 & 0.9612 \\
& & 3D & 0.2515 & 0.0812 & 0.2850 & 0.9479 \\
\hline
\end{tabular}
\end{table}

From Table~\ref{tab:structure_ablation_results}, we observe that the \textbf{No Mask} group required the highest number of epochs (402) to converge, significantly more than other groups. This confirms the effectiveness of the sparse self-attention mask in improving training efficiency. The improvement is likely due to the mask dynamically filtering redundant positional correlations, thus reducing computational complexity, as supported by Child et al.~\cite{45}. Furthermore, the higher errors in each dimension compared to the baseline indicate that the sparse mask enhances the model’s ability to capture local correlations between joint angles and end-effector coordinates by focusing attention on key regions.

The \textbf{Body Mask} group also showed a higher error across all axes compared to the baseline, indicating the superiority of our proposed mask design, which better emphasizes the geometric relationships between robotic arm links and mitigates global noise interference.

The \textbf{No ResNet} group produced the highest errors among all groups, surpassing both the Baseline and the \textbf{No Transformer} group. This result suggests that the Body Transformer encoder alone cannot achieve optimal performance in prediction tasks; it must be coupled with residual blocks in the prediction head. Residual connections preserve low-level kinematic features, while the Transformer excels at capturing high-level spatio-temporal dependencies. Their combination significantly improves the accuracy of inverse kinematics.

The \textbf{No Transformer} group also showed higher errors in all dimensions, with 3D MAE increasing by 18\% relative to the baseline. This shows that relying solely on ResNet is insufficient for high-precision predictions. While ResNet can extract single-frame features, the multihead self-attention in Transformer enables modeling of inter-frame angular dependencies.

In conclusion, the BoTER model achieves high-precision mapping from six-axis joint angles to end-effector coordinates through synergistic integration of sparse attention masks (for localized focus), BoT encoders (for long-range dependency modeling) and ResNet blocks (for hierarchical feature retention). These results validate the effectiveness of hybrid architecture in the prediction of robotic kinematics.

\section{Inverse Joint Angle Compensation Verification on UR5 Industrial Robot}

After training with 724 samples, the optimal high-precision SPI-BoTER model was obtained for forward prediction of industrial robot end-effector positions. In the next step, we validate the effectiveness of the overall framework for inverse joint angle error compensation on the UR5 industrial robot using a gradient descent method.

\begin{figure}[h!]
\centering
\includegraphics[width=0.75\textwidth]{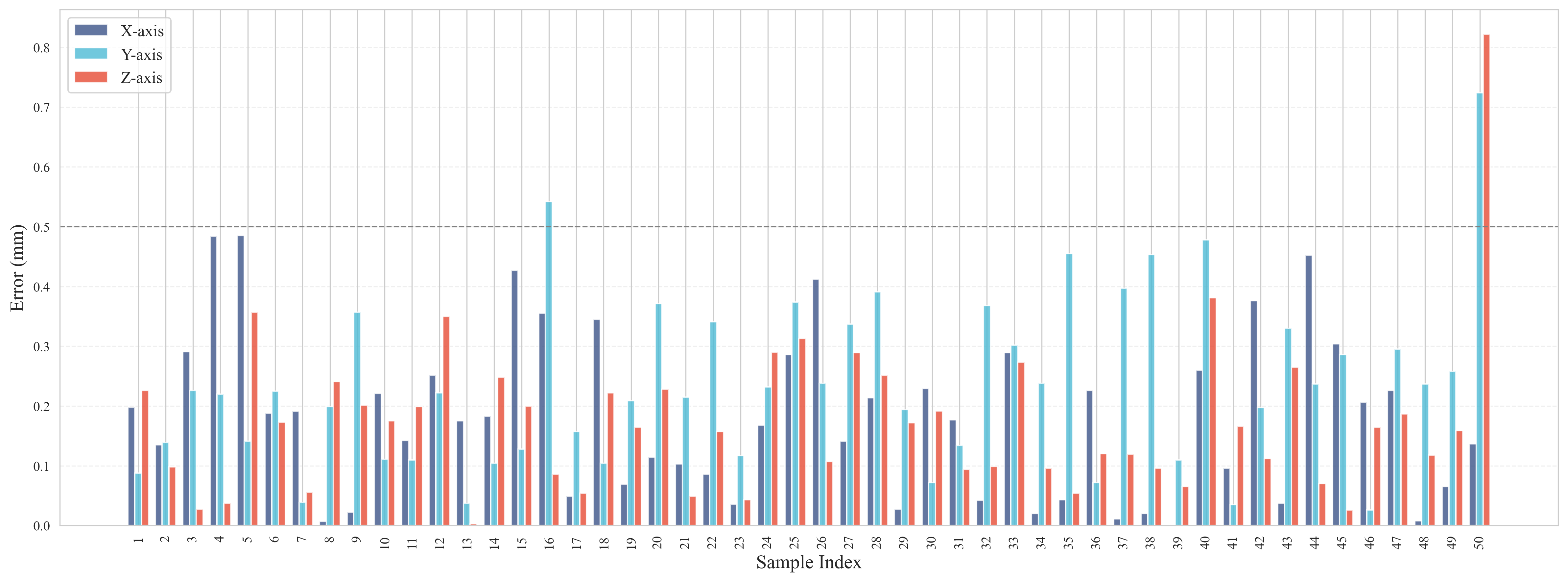}
\caption{Position Error of 50 Randomly Sampled Test Points}
\label{fig:random_position_error}
\end{figure}

\begin{table}[h!]
\centering
\caption{Inverse Compensation Results on UR5 Robot}
\label{tab:inverse_compensation}
\begin{tabular}{lccc}
\hline
\textbf{Axis} & \textbf{Standard Deviation (mm)} & \textbf{Maximum Error (mm)} & \textbf{Minimum Error (mm)} \\
\hline
X & 0.1355 & 0.4850 & 0.0010 \\
Y & 0.1438 & 0.7240 & 0.0260 \\
Z & 0.1314 & 0.8220 & 0.0030 \\
\hline
\end{tabular}
\end{table}

In this experiment, a random seed of 18 was used to select 50 theoretical position targets from the UR5 dataset test set. The corresponding joint angles were taken as the initial input, denoted as $\boldsymbol{\theta}_{\text{initial-deg}}$. Based on the trained SPI-BoTER model, angle compensation $\Delta \boldsymbol{\theta}$ was optimized using gradient descent to minimize the distance to the target positions. The compensated joint angles were then fed to the robot controller to drive the UR5 industrial robot, and the actual end-effector positions were recorded using the TrackScan Sharp system.

The effectiveness of the compensation strategy was evaluated by the maximum deviation between the measured and target positions. As shown in Fig.~\ref{fig:random_position_error} and Table~\ref{tab:inverse_compensation}, the maximum absolute deviation was \textbf{0.62 mm}, with a maximum standard deviation of only \textbf{0.1495 mm}. Under a tolerance requirement of $\pm0.5$ mm, only 2 out of 50 test points failed to meet the threshold, yielding a success rate of \textbf{98\%}. These results fully demonstrate the high accuracy of the SPI-BoTER model in error prediction and the feasibility of the proposed compensation strategy for real-world industrial applications.

\section{Conclusion}

This study proposes the SPI-BoTER framework, which achieves a significant advancement in industrial robot end-effector position prediction and error compensation through a dual-branch mechanism-data collaborative architecture. A sparse self-attention mask tailored to six-axis kinematics enables deep fusion between the global modeling capabilities of Transformer and structural constraints. This design reduces parameter count by 34\% while achieving a 3D mean absolute error of \textbf{0.2515 mm} on the UR5 dataset, improving accuracy by \textbf{35.16\%} compared to traditional DNN methods.

The hybrid spatial-physics loss function aligns predictions with the topological structure of the theoretical model through a Euclidean distance matrix alignment mechanism, increasing spatial similarity to \textbf{97.3\%} and effectively suppressing extreme errors under challenging working conditions. The inverse compensation algorithm, coupled with gradient descent optimization, converged within 147 iterations to an accuracy of \textbf{0.01 mm}, offering a viable solution for real-time closed-loop control.

However, several limitations remain. First, the experimental dataset is limited to static and randomly sampled points, lacking validation under continuous trajectory motion. Second, the sparse attention mask is specifically designed based on the UR5 robot's structure, and its generalizability to heterogeneous multi-joint robots requires further investigation. Given the excellent temporal modeling capabilities of Transformers, we aim to extend SPI-BoTER to the time dimension in future work, which may further enhance its deployment performance in real-world robotic applications.
\section*{Acknowledgement}

This work was supported by the National Natural Science Foundation of China (Grant No. 52335001).

\end{document}